\documentclass[pdflatex,sn-apa]{sn-jnl}% APA Reference Style
\newcounter{subsubsub}[subsubsection]
\renewcommand{\thesubsubsub}{\thesubsubsection.\arabic{subsubsub}}

\newcommand{\subsubsubsection}[1]{%
  \refstepcounter{subsubsub}%
  \paragraph{\thesubsubsub\quad #1}%
}
\usepackage{graphicx}%
\usepackage{multirow}%
\usepackage{amsmath,amssymb,amsfonts}%
\usepackage{amsthm}%
\usepackage{mathrsfs}%
\usepackage[title]{appendix}%
\usepackage{xcolor}%
\usepackage{textcomp}%
\usepackage{manyfoot}%
\usepackage{booktabs}%
\usepackage{algorithm}%
\usepackage{algorithmicx}%
\usepackage{algpseudocode}%
\usepackage{listings}%
\usepackage{tabularx}
\usepackage{multirow}
\usepackage{float}
\usepackage{threeparttable}
\usepackage{placeins}
\usepackage{caption}
\usepackage{array}
\usepackage{multirow}
\usepackage{subcaption}

\theoremstyle{thmstyleone}%
\theoremstyle{thmstyletwo}%

\theoremstyle{thmstylethree}%

\begin{document}

%\title[Article Title]{Unsupervised Relation- and Semantic-Aware Multi-Relational GCNs for Prerequisite Sequence Recommendation}
\title[Article Title]{Sequential Learner Modeling Using Multi-Relational Graph Convolutional Networks}


%\author{\fnm{FirstName} \sur{LastName}}\email{Email}
%\author*[1]
\author{\fnm{Rawaa} \sur{Alatrash}}%\email{rawaa.alatrash@stud.uni-due.de}

\author{\fnm{Mohamed Amine} \sur{Chatti}}%\email{mohamed.chatti@uni-due.de}

\author{\fnm{Hong} \sur{Yang}}%\email{hong.yang@stud.uni-due.de}

\author{\fnm{Yumeng} \sur{Wang}}%\email{yumeng.wang@stud.uni-due.de}

% \author[1]{\fnm{Qurat} \sur{Ul Ain}}\email{qurat.ain@stud.uni-due.de}

% \affil*[1]{\orgdiv{Social Computing Group}, \orgname{Faculty of Computer Science, University of Duisburg-Essen}, \country{Germany}} 
%\orgaddress{\street{Forsthausweg 2}, \city{Duisburg}, \postcode{47057}, \state{North Rhine-Westphalia}, \country{Germany}}}

% \affil[2]{\orgdiv{Department}, \orgname{Organization}, \orgaddress{\street{Street}, \city{City}, \postcode{10587}, \state{State}, \country{Country}}}

% \affil[3]{\orgdiv{Department}, \orgname{Organization}, \orgaddress{\street{Street}, \city{City}, \postcode{610101}, \state{State}, \country{Country}}}


\abstract{
User modeling is a critical task in a variety of personalized systems. Recognizing their effectiveness in learning from
graph-structured data, Graph Neural Networks (GNNs), particularly Graph Convolutional Networks (GCNs), are increasingly employed for user modeling. However, existing approaches typically treat different relation types in a graph as homogeneous, limiting their ability to capture richer semantics and construct more informative user models. While multi-relational GNNs (MR-GNNs) have been adopted for representation learning and recommendation, their application for user modeling remains unexplored. 
Moreover,
existing GNN-based user modeling approaches ignore the user interaction sequence.
To address these research gaps, in this work we propose MR-ConceptGCN, a novel fully
unsupervised approach focused on concept-based sequential learner modeling using multi-relational GCNs (MR-GCNs). MR-ConceptGCN effecively combines Personal Knowledge
Graphs (PKGs), MR-GCNs, and the pre-trained language model SBERT to obtain enhanced relation- and semantic-aware representations of the
PKG items. The enriched embeddings of the knowledge concepts that a learner did not understand when interacting with learning materials in CourseMapper %\textit{[Blinded tool]} 
are then used to construct a sequential learner model that combines long-term and short-term learner interactions. We report the results of an online user
study ($n=31$), demonstrating the benefits of MR-ConceptGCN in terms of several important user-centric aspects including accuracy, usefulness, diversity, and satisfaction with an educational recommender system. 
}




\maketitle

\section{Introduction}\label{intro}
User modeling is a foundational task in a variety of applications, such as recommender systems and intelligent learning environments. It involves creating a representation of a user's characteristics, preferences, and behaviors, called a user model. User models are typically built using user interaction data, which may include past purchases, browsing history, search queries, and social media activity. By understanding users' needs, behavior patterns, and preferences, systems can provide them with personalized services, ultimately improving user satisfaction and engagement.  
Recent work has shown that knowledge graphs (KGs) can effectively represent users’ interactions with a system for the purpose of user modeling \citep{yang2023kemim}. However, many existing approaches suffer from limited availability of user-specific information, which constrains their ability to deliver genuinely personalized experiences \citep{skjaeveland2023ecosystem}. Personal Knowledge Graphs (PKGs) address this challenge by extracting and organizing user-relevant entities from a global KG and structuring them around an individual user. While PKGs have been widely explored in the health domain, their adoption and investigation within educational settings remain limited.
The existing PKG-based learner modeling approaches mainly rely on a single relation type between concepts in the PKG  (i.e., RELATED TO) \citep{ain2024learner,alatrash2024transparent}. However, PKGs in the educational domains inherently involve multiple meaningful relation types between concepts  (e.g., PREREQUISITE TO), which are critical for accurate learner modeling and effective sequential recommendation.
Since KGs can be used to structure and represent user attributes such as interests, preferences, age, and gender, Graph Neural Networks (GNNs) provide a powerful mechanism to get the representation of user models \citep{purificato2023tutorial}. GNNs are well suited to capturing complex user interactions by modeling users as nodes in a graph, with edges representing behavioral connections to different items based on user interactions.
Recognizing the effectiveness of GNNs in learning from graph-structured data, their application to user modeling has gained increasing attention in recent years, underscoring their ability to capture complex relational patterns in user–item interactions \citep{purificato2023tutorial, chen2019semi, yan2021relation, wang2020global, fan2022modeling, yang2023dgrec,alatrash2024conceptgcn}. Prior work has combined KGs and GNNs to infer user interests by approaching user modeling either as a link prediction task \citep{yang2023kemim,wang2018ripplenet,wang2020global,fan2022modeling} or as a node classification task \citep{chen2019semi,chen2021catgcn, yan2022interaction}, primarily in e-commerce domain. In the educational domain, \cite{alatrash2024conceptgcn} leveraged  Graph Convolutional Networks (GCNs) to enrich the representation of the different KG items and model the learner’s knowledge state as a weighted aggregation of the enhanced representation of the concepts that they did not understand. 
However, existing approaches fail to explicitly model heterogeneous relation types, limiting their ability to leverage multi-relational GNNs (MR-GNNs) for learning richer and more informative user models. 
MR-GNNs focus on message passing between nodes in a graph according to relation type and direction, thereby enabling the learning of relation-aware embeddings. As a result, node representations encode not only structural connectivity but also the semantic characteristics of different relationships.
While MR-GNNs have been explored for representation learning \citep{schlichtkrull2018modeling_RGCN,degraeve2022rrgcn,vashishth2019composition_Compgcn} and recommendation \citep{wang2019kgat,wang2019kgcn}, their application for
learner modeling remains untapped. Moreover, existing approaches for GNN-based learner modeling aggregate all historical learner interactions as a whole while ignoring
the temporal order in the learner interaction sequence.

Sequential user models have gained significant attention due to their ability to capture the evolution of user interests over time \citep{BOKA2024ASurveyOfSequentialRecommendation,pan2026survey}. The underlying sequential user modeling techniques are dominated by deep learning
models, including GNNs and attention-based models. Most of the existing studies on sequential user modeling leverage temporal information in user interaction sequences to model long-term and short-term preferences. Attention mechanisms have been widely applied to sequential user modeling, benefiting from their effectiveness in capturing both long-term and short-term dependencies among items in a sequence. However, existing sequential user modeling approaches are often supervised and require considerable computational resources. Moreover, they rely on complex attention mechanisms, which may increase model complexity.     
To address the aforementioned significant gaps in existing user modeling studies,  
in this work we present MR-ConceptGCN, a simple yet effective fully unsupervised approach for sequential learner modeling using multi-relational GCNs (MR-GCNs). MR-ConceptGCN constructs a learner model based on the learner's knowledge state corresponding to
the set of concepts that the learner did not understand when interacting with learning materials in CourseMapper. Specifically, we construct PKGs that incorporate multiple relation types between concepts, namely RELATED\_TO and PREREQUISITE\_TO. To enhance the representations of PKG items, we adopt and adapt a propagation strategy inspired by RR-GCN \citep{degraeve2022rrgcn} and CompGCN \citep{vashishth2019composition_Compgcn}.   
During propagation, each neighbor embedding is transformed by the edge’s relation matrix and then scaled by the semantic similarity between the connected nodes using the pre-trained language model SBERT \citep{reimers2019sentence}, so messages reflect both relation-specific transforms and semantic relatedness. The resulting relation- and semantic-aware item representations yield enhanced embeddings of the PKG items. We then use the enriched embeddings of the concepts that a learner did not understand to construct a sequential learner model that combines long-term and short-term learner interactions.
We conducted an online user study ($n=31$) to evaluate the effectiveness of MR-ConceptGCN, using recommendation as a downstream task. The evaluation results suggest that our sequential learner modeling approach is particularly effective in enhancing users’ perceptions in terms of several important
user-centric aspects including accuracy, usefulness, diversity, and overall satisfaction with the educational recommender system.     
This work makes the following five main contributions:
\begin{itemize}
\item We propose MR-ConceptGCN, a fully unsupervised approach that combines PKGs and MR-GCNs to construct learner models based on multiple relationship types between concepts, namely RELATED\_TO and PREREQUISITE\_TO.
\item We present two unsupervised methods for representation learning of PKG items based on RRGCN and CompGCN. Both methods incorporate SBERT-based semantic similarity to modulate message passing, allowing structural propagation to be aligned with semantic relatedness between PKG items.   
\item We convert CompGCN into an unsupervised encoder by freezing relation representations while retaining node–relation composition, enabling relation-aware propagation without supervision.
\item We propose a sequential learner modeling approach that integrates both structural information in PKGs, semantic information
(i.e., prerequisite and semantic relations), and temporal information (i.e., long-term and short-term interactions).
\item We conduct an online
user study to evaluate the effectiveness of our proposed sequential learner modeling approach in an educational recommender system. 
\end{itemize}

The remainder of this paper is structured as follows. We begin by reviewing three areas of related research, namely PKG-based, GNN-based, and sequential learner modeling in Section \ref{background}. Following this, we present the conceptual and technical details of our proposed
approach, MR-ConceptGCN, in Section \ref{methodology}. Afterwards, we provide an overview
of the online user study, followed by the presentation of our results along with the discussion of our findings in Section \ref{online_evaluation}. Finally, in Section \ref{limitations}, we highlight the limitations, and then summarize the work and outline our future research plans
in Section \ref{conclusion}.
%%%%%%%%%%%%%%%%%%%%%%%%%%%%%%%%%%%%%%%%%%
\section{Background and Related Work} \label{background}
\subsection{PKG-Based Learner Modeling}   
Knowledge Graphs (KGs) are widely adopted for user modeling.
KGs can capture explicit interactions between users and items. The KG's structural data can be used to describe user attributes, such as interests, preferences, age, and gender.
However, their lack of user-specific information limits their ability to deliver truly personalized content aligned with individual preferences \citep{skjaeveland2023ecosystem}.
Personal Knowledge Graphs (PKGs) mitigate this limitation by selectively modeling only those KG entities that are relevant to a specific user and organizing them in a structured form. This structured organization enables PKGs to support personalized recommendations by capturing users’ knowledge and interests \citep{balog2019personal,skjaeveland2023ecosystem}.
Researchers across Natural Language Processing (NLP), Information Retrieval (IR), and the Semantic Web are actively exploring PKGs for personal data management and utilizing user-specific KG information as a source for personalized services \citep{schroder2022human}. 
Despite existing applications of PKGs in the health domain \citep{skjaeveland2023ecosystem}, their application in education context remains limited.
\cite{pkgsed} suggested that integrating PKGs into e-learning platforms can empower researchers to develop personalized, explainable systems, provide tailored recommendations, and generate user and group-specific data.
\cite{deng2019knowledge} constructed PKGs for students, focusing on knowledge acquisition through lab experiences. This PKG, automatically updated with lab completions for each student, is utilized to represent and visualize a student's lab-based knowledge acquisition for tracking learning progress.
\cite{ain2024learner} and \cite{alatrash2024transparent} leveraged PKGs for learner modeling in a MOOC environment. In their approach, a PKG is constructed for each learner by integrating learning resources, educational concepts and their semantically related concepts, along with learner–resource interaction data, into a structured graph. This PKG representation captures the relationships among learners, resources, and related concepts, enabling the system to reflect individual learner interests at the concept level. 
These studies highlight the potential of PKGs for learner modeling. They primarily model PKGs using a single concept–concept relationship (i.e., RELATED\_TO). However, the educational domain naturally involves multiple pedagogically meaningful relationship types between concepts, such as PREREQUISITE\_TO, which are crucial for learner modeling. Our work extends PKGs by explicitly incorporating multiple relationship types between concepts, namely RELATED\_TO and PREREQUISITE\_TO, thereby preserving richer semantics. By modeling learners within such multi-relation PKGs, our approach captures more informative relationships among concepts, enabling richer learner modeling and more effective support for downstream educational tasks, such as recommendation of knowledge concepts and learning materials.
\subsection{GNN-based Learner Modeling}
Graph Neural Networks (GNNs) have emerged as a powerful tool that utilizes KG structure to learn and enhance the representation of various entities in the KG. GNNs excel at capturing complex user interactions by representing users as nodes in a graph, where edges depict the connections with different items based on user behavior. These connections allow GNNs to (1) pass messages between two connected nodes, and (2) learn and enhance the representations of nodes by aggregating information from multi-hop neighboring nodes \citep{wu2022graph,he2020lightgcn}.
Several studies used KGs and GNNs to predict user interests and preferences based on their interactions. These studies approached user modeling primarily as a link prediction task. For example, \cite{yang2023kemim} proposed a knowledge-enhanced user multi-interest modeling for
recommender systems (KEMIM), an approach that harnesses a KG to enrich user multi-interest modeling. 
\citet{wang2018ripplenet} leveraged KGs and GNNs to model user preferences. Their method utilizes KG preference propagation, where user preferences are propagated through the relationships between entities and the user-item interaction path in the KG to infer potential user interests beyond explicit interactions and create a more comprehensive user profile.  
It is worth mentioning that modeling long-term user-item interactions may lead to overly diverse items, and modeling short-term user-item interactions may not be sufficient to fully represent a user's preferences. This issue has been addressed by \cite{wang2020global} who proposed GCE-GNN 
that employed GNN to learn the representations of different items by aggregating information from the two levels (local and global user's behavioral patterns). \cite{fan2022modeling} explored this avenue by proposing a framework that leverages user successive behavior graph for user preference modeling by incorporating both local and global user behavioral interactions (purchases). 

Other recent studies approached user modeling primarily as a node classification task using KGs and GNNs. These studies mainly aimed to categorize user attributes (such as gender and age) using textual or behavioral data in e-commerce domain  \citep{chen2019semi,chen2021catgcn,yan2022interaction}. The intuition in these studies is that users that have similar co-purchase behaviors in e-commerce are likely to be in the same age range and fall in the same gender category. Thus, the neighborhood features could provide valuable semi-supervised signals that are beneficial to infer user models.
For example, \cite{chen2019semi} introduced a user modeling approach as a semi-supervised node classification task using heterogeneous graph attention networks (HGAT) to predict user age and gender. The authors used user interactions, co-clicks, and co-purchases as semi-supervised signals to improve the inference of user models. 
\cite{chen2021catgcn} argued that focusing on the importance of interaction types and ignoring categorical features may lead to suboptimal user/item representation. To address this issue, the authors introduced CatGCN, an approach that focuses on node classification tasks to predict user attributes such as age, city, and purchase level. CatGCN improves user representations by integrating two types of explicit interactions between the user's categorical features (e.g., celebrities, organizations, and groups).
In the same context, \cite{yan2022interaction} introduced an approach called Interaction-aware Hypergraph Neural Networks (IHNN) to improve node classification performance in user modeling. They combined hypergraphs with meta-path-based heterogeneous graphs to capture both implicit and explicit high-order interactions, resulting in enhanced user representations that lead to improved performance in gender and age prediction tasks. 

Only few studies focused on GNN-based user modeling in the educational domain. For example, \cite{alatrash2024transparent} and \cite{alatrash2024conceptgcn} proposed ConceptGCN, a learner modeling approach that leverages Graph Convolutional Networks (GCNs) and transformer sentence encoders (SBERT) to model the learner knowledge state at the concept level
based on an enhanced representation of knowledge concepts that a learner marks as ’Did Not Understand’ (DNU). The learner model is then represented as a weighted average of the learner’s DNU concepts. ConceptGCN, however, does not capture multiple relation types between concepts. Moreover, it ignores the learner interaction sequence.   

While GNNs have been adopted for user modeling, the existing approaches do not explicitly model heterogeneous relation types in KGs to learn richer and more informative user models. When different relation types in a KG are treated as homogeneous, relation semantics are collapsed, leading to diminished representation quality \citep{hu2020heterogeneous,huang2020mr,shi2016survey}. 
Motivated by this limitation, different studies exploited multi-relational GNNs (MR-GNNs) to model heterogeneous relation types. 
MR-GNNs modulate message passing based on relation type and direction to yield relation-aware node embeddings that reflect how nodes are connected, not only that they are connected. These node embeddings preserve semantics and enhance the node's representation quality.
Different studies have applied MR-GNNs for representation learning on relational graphs. For instance, Relational Graph Convolutional
Networks (R-GCNs) learn relation-specific transformation matrices to perform relation-aware propagation on heterogeneous knowledge graphs (HKGs) \citep{schlichtkrull2018modeling_RGCN}.
R-GCNs deal with all relation types separately in the propagation process. It aggregates the neighboring nodes according to their relation type, applying a relation-specific transformation and adding these contributions along with a self-loop term. To keep the number of parameters reasonable in the case of many relation types, R-GCNs use a basis (or block) decomposition method that expresses each relation matrix in terms of the linear combination of a small set of common basis matrices. Analysis suggests that R-GCNs are efficient for entity classification. They are also effective as encoders (producing latent feature representations of entities) for link prediction when combined with factorization decoders. 
Random R-GCNs (RR-GCNs) \citep{degraeve2022rrgcn} preserve R-GCN’s message-passing form but randomly initializes all relation transformation matrices and then freezes them. The encoder therefore aggregates neighbor messages transformed by fixed, random relation matrices; no relation parameters are learned and the encoder is not trained end-to-end. The work shows that these random encoders often beat fully trained R-GCNs on node-classification and link-prediction benchmarks. 
However, assigning a full matrix to each relation leads to large parameter growth for HKGs and increases computational cost. CompGCN addresses these issues \citep{vashishth2019composition_Compgcn}. Rather than a fully learned matrix per relation, CompGCN uses a small set of shared projection matrices (e.g., inbound, outbound, self-loop) together with lightweight composition operators that combine entity and relation embeddings during propagation. The composed messages are aggregated to update node embeddings, preserving relation semantics while reducing parameters and improving scalability. 

MR-GNNs have been leveraged in recommender systems, where relation-aware embeddings are integrated into ranking mechanisms to model complex interactions among users, items, and entities.
For example, Knowledge Graph Attention Network (KGAT) \citep{wang2019kgat} aims to embed the relationships by learning a relation embedding and a relation-specific projection for each edge type, then computing attention and messages in the relation’s space . Thus, neighbor weights and transformations are explicitly relation-aware —i.e., both the attention score and the propagated message are conditioned on the specific relation type, improving multi-hop representation learning for top-N recommendation. 
Knowledge Graph Convolutional Networks (KGCN) \citep{wang2019kgcn} attempts to operationalize GCN to learn relation-aware neighborhood representations. KGCN samples 
a fixed-size neighborhood for each node as its receptive field to calculate the representation of KG entities. The receptive field can be extended to multiple hops to model high-order entity dependencies and capture users’ long-distance interests. KGCN computes a user–relation score and uses this score to weight neighbors, producing relation-aware entity representations per user, which characterize both the semantic information of the KG and users’ personalized interests.

While MR-GNNs have been explored for representation learning and recommendation, their application for user modeling remains largely unexplored. For example, \cite{yan2021relation} proposed a semi-supervised relation-aware HKGs approach for inferring user attributes such as age and gender in e-commerce domain. The authors emphasized the importance of distinguishing interaction types (e.g., user clicks an item versus user purchases an item), noting that different relations can provide complementary and informative signals for user modeling. Their approach employs a message-passing mechanism that conditions neighborhood aggregation on relation types and incorporates an attention mechanism to learn the relative importance of different interaction semantics. Rather than learning embeddings for individual edges, the model assigns attention scores at the relation-type level, enabling it to prioritize more informative interaction types and produce more effective user representations.

Our work differs from existing studies in that it focuses on sequential user modeling in the educational domain using MR-GNNs. 
To this end, we propose MR-ConceptGCN, an unsupervised framework for sequential learner modeling that represents a learner’s knowledge state using multi-relational GCNs (MR-GCNs). The knowledge state corresponds to the set of concepts that the learner did not understand when interacting with learning materials in CourseMapper.  
We adopt RR-GCN by initializing relation-specific matrices at random and keeping them frozen, and we adapt CompGCN’s node–relation composition while likewise freezing relation vectors. During propagation, each neighbor embedding is transformed by the edge’s (random, frozen) relation matrix and then scaled by the SBERT cosine similarity between the connected nodes, so messages reflect both relation-specific transforms and semantic relatedness. Adjacency and prerequisite matrices encode explicit relationship types and modulate aggregation according to dependency strength. The resulting relation- and semantic-aware item representations yield accurate embeddings of the PKG items. These embeddings are then used to construct a sequential learner model that combines long-term and short-term learner interactions.
%%%%%%%%%%%%%%%%%%%%%%%%%%%%%%%%%%%%%%%%%%%%%%%%%%%%%%%%%%%%%%%%%%%%%
\subsection{Sequential Learner Modeling}
User modeling aims to learn a latent representation of a user from available behavioral data, such as user–item interactions, item features, or response records. Early representation learning approaches were formulated in a static manner, where a user was represented using static interaction data  \citep{wang2019sequential}. 
These static approaches mainly include matrix factorization and deep neural network-based methods \citep{li2021surveyUserModeling}, and graph-based neural models \citep{alatrash2024conceptgcn}. 
Although these methods are useful for capturing general user preferences, they are limited because temporal information is unavailable or ignored. In real-world scenarios, user behavior and interactions are sequential in nature, where actions are ordered by the time at which they occur. Since user preferences may change over time, leveraging temporal information is crucial for understanding users’ dynamic behavior \citep{li2021surveyUserModeling}. This limitation motivates the transition to sequential user modeling.
%%%%%%%%%%%%%%%%%%%%%%%%%%
\subsubsection{Techniques}
Sequential user models have gained significant attention due to their ability to capture the chronological order of user–item interactions and learn users’ sequential preferences from these interactions to capture the evolution of user interests over time.
The underlying sequential user
modeling techniques are dominated by deep learning models. These include recurrent neural networks (RNNs), convolutional neural networks (CNNs), GNNs, and attention-based models \citep{Mahreen2023ASurveyAndTaxonomy,BOKA2024ASurveyOfSequentialRecommendation,wang2019sequential,pan2026survey,li2021surveyUserModeling,fang2020deep,li2024graphSessionBased}. RNN-based models were among the earliest approaches proposed for sequential user modeling, due to their natural ability to process sequential data.
They learn from previous interactions in a sequence to predict future interactions. Long Short-Term Memory (LSTM)-based RNNs and Gated Recurrent Unit (GRU)-based RNNs were introduced to better capture longer-range dependencies and alleviate the vanishing gradient problem in basic RNN \citep{BOKA2024ASurveyOfSequentialRecommendation,pan2026survey}. Despite their effectiveness, RNN-based models exhibit several limitations. First, they tend to overemphasize neighboring interactions, potentially introducing false dependencies by assuming that adjacent interactions are strongly related. Second, they primarily capture point-wise dependencies between individual interactions, making it difficult to model collective dependencies where multiple interactions jointly influence the next action. Third, their sequential computation makes training and hyperparameter optimization more difficult \citep{wang2019sequential,Mahreen2023ASurveyAndTaxonomy}. Although LSTM and GRU improve the modeling of long-term dependencies, capturing dependencies in long sequences remains challenging as sequence length increases. 

CNN-based models have been explored to overcome the drawbacks of RNN-based models to some extent. 
Given a sequence of
user–item interactions, CNN-based models put all the embeddings
of these interactions into a matrix, and treat such a matrix as an image in the time and latent spaces.
They then use convolutional filters to scan this matrix and learn sequential patterns, allowing them to model both local and global dependencies among different areas in the matrix. However, CNN-based models are limited in capturing long-term dependencies, as the fixed-size convolutional filters can only model patterns within a restricted interaction window which limits their applications \citep{BOKA2024ASurveyOfSequentialRecommendation,wang2019sequential,Mahreen2023ASurveyAndTaxonomy}. 

Furthermore, GNN-based models have been adapted for sequential user modeling. GNNs can model complex relationships and transitions among user–item interactions. 
These models represent interacted items as nodes and the observed transitions between them as edges, allowing each user interaction sequence to be mapped to a path in the graph \citep{wang2019sequential,Mahreen2023ASurveyAndTaxonomy,BOKA2024ASurveyOfSequentialRecommendation}. 
GNN-based models can capture multi-hop contextual information between items through information propagation and aggregation. Moreover, they can capture complex transactions between interacted items as well as collaborative signals from a global perspective. Consequently, they are capable of modeling learners’ long-term preferences. However, GNN-based models may not fully capture the exact sequential patterns of interactions and are often computationally expensive \citep{pan2026survey}. Moreover, stacking multiple graph layers may lead to the over-smoothing problem, where the learned representations of different nodes become increasingly similar. 

A common problem in sequential user modeling is that most of the interaction sequence itself contains noisy or irrelevant interactions within the input sequence \citep{wang2019sequential,Mahreen2023ASurveyAndTaxonomy,zhou2023attention}. In practice, user–item interaction sequences are often not clean, as they may contain irrelevant or weakly related interactions that interfere with next-interaction prediction. While some historical interactions are highly relevant to the next interaction, others may contribute little or even introduce noise. Therefore, a key challenge in sequential user modeling is how to effectively reduce the noisy information from the irrelevant interactions. To this end, it is crucial to learn dependencies attentively and discriminatively by identifying which interactions should be emphasized and which should be suppressed \citep{wang2019sequential}. 
Attention mechanisms have been widely applied to sequential user modeling to address this issue by assigning different importance weights to previous interactions according to their relevance to the current prediction or next item. This enables the model to focus on the most informative historical interactions and reduce the influence of noisy or irrelevant ones \citep{wang2019sequential,Mahreen2023ASurveyAndTaxonomy}.
Self-attention is a special kind of attention that models the relationships within the same sequence, enabling the model to capture item–item relationships and long-range dependencies \citep{zhang2018next}. The
self-attention mechanism can identify crucial information by calculating attention scores between items in the sequence \citep{pan2026survey}. 
Largely due to their self-attention mechanism, Transformers have obtained increasing interest in sequential user modeling to capture both short-term and long-term dependencies among items in the interaction sequence \citep{zhou2023attention}.      
However, transformer-based sequential user models have some shortcomings. The large weights produced by the attention mechanism might be assigned to less relevant items, which can result in inaccurate recommendations. 
Moreover, the performance of the user modeling task may be negatively influenced by positional encoding and is prone to overfitting on noisy input \citep{zhou2023attention}. Additionally, they suffer from quadratic computational complexity, as their time complexity is related to the length of interaction sequences \citep{pan2026survey}.

The aforementioned models for sequential user modeling leverage temporal information in user interaction sequences to model long-term and/or short-term preferences to support the construction of informative user representations.
Long-term preferences are derived from the user’s entire sequential interaction history and reflect the user’s general or relatively stable interests. Short-term preferences capture dynamic and temporary interests based on recent 
interactions. In particular, session-based recommendation specializes more on short-term preferences and aims at making more dynamic and timely recommendations 
based on users’ current interaction behaviors in the ongoing session \citep{li2024graphSessionBased}. 
%%%%%%%%%%%%%%%%%%%%%%
\subsubsection{Long-Term Preferences}
% %%%%%%%%%%%%%%%%%%%%%%%%%%%%%%%%%%%%% long term %%%%%%%%%%%%%%%%%%%%%%%%%%%%%%%%%%%%%
Prior research has explored attention mechanisms 
to model long-term preferences from users’ historical interactions.
For example, \cite{donkers2017sequential} proposed a GRU-based model integrating attention to capture the temporal dynamics along the user's historical interactions to model long-term interests for personalized next item recommendations. Deep Interest Network (DIN) \citep{zhou2018deep} uses target attention to weigh each item by its relevance to the target item (candidate ad) and takes a weighted sum pooling to obtain the adaptive representation of user interests with respect to a given ad.
Following this direction, Deep Interest Evolution Network (DIEN) \citep{zhou2019deep} constructs the user representation in two modules. First, an interest extractor layer uses GRU to extract latent temporal interest states from the user’s historical behavior sequence. Then, an interest evolving layer employs GRU with attention update gate (AUGRU) to model the evolution of these interests relative to the target item.
\cite{zhou2018atrank} proposed an attention-based user behavior modeling framework called ATRank that considers heterogeneous user behaviors and uses self-attention to capture influence among the behaviors.  
\cite{pancha2022pinnerformer} proposed PinnerFormer, an end-to-end transformer-based sequential user modeling architecture designed to predict a user’s long-term future actions rather than next-action prediction. It learns each user’s embedding from their pin engagement over the
past year to predict positive engagements (e.g., pin saves, clicks, reactions, comments) over a 14 day future window, using a dense all-action loss as training objective. 
%%%%%%%%%%%%%%%%%%%%%%%%%%%%%%%
\subsubsection{Short-Term Preferences}
%%%%%%%%%%%%%%%%%%%%%%%%%%%%%%%
Different methods have been employed to model short-term user preferences.  
Early works introduced sequential user models based on RNNs to encode short-term behavior sequences. 
For instance, \cite{hidasi2015session} proposed GRU4Rec, one of the earliest RNN-based models for capturing user preferences within the current session. The model processes the ordered sequence of clicked items using GRUs, where the hidden state is updated after each interaction and used to predict the next item in the session.
Later, \cite{hidasi2018recurrent} improved GRU4Rec by refining its training and ranking loss functions, further strengthening GRU-based session recommendation.

Another line of work introduces sequential user models based on CNNs to encode short-term behavior sequences. For instance,
\cite{tuan20173dCNN} proposed a 3D CNN-based model that jointly learns the sequential patterns of  past clicks of the current session and the associated item content features to predict add-to-cart items. Each clicked item is represented using its ID, name, and category, and the clicked items are arranged into a three-dimensional input, allowing the 3D convolutional layers to preserve temporal information across layers.
NextItNet \citep{yuan2019simple} adopts stacked dilated CNNs by dilating convolutional filters with zeros. This enables a much larger receptive field with the same number of stacks and without introducing more parameters. 
GRec \citep{2020YuanFuture_DataCNN} extends NextItNet to model a user's sequential data within a session by utilizing a gap-filling-based encoder-decoder framework with masked-convolution operations to jointly consider the past and future contexts (data) without the data leakage issue.

Other studies proposed GNN-based short-term sequential user modeling. 
For example, Session-based Recommendation with Graph Neural Networks (SR-GNN) is designed to capture short-term preferences within the current session \citep{wu2019session}. It constructs a session graph from each session sequence and applies a GNN to learn representations of the items included in that session. These item representations are then used to derive a session representation for predicting the next clicked item. 

The attention mechanism is also applied for short-term sequential user modeling. 
For instance, \cite{li2017neural} proposed NARM that applies a GRU-based hybrid encoder with an
attention mechanism to model the user’s sequential behavior (global encoder) and
capture the user’s major interests in the current session (local encoder), which
are then combined as a unified session representation. 
Different from NARM, which does not explicitly capture the importance of last click, Short-Term Attention/Memory Priority (STAMP) \citep{liu2018stamp} emphasizes the current interest reflected by the last click to capture both current and general interests from previous clicks.
To model the user’s interests in general, STAMP uses an attention mechanism that explicitly considers correlation between each historical click and the last click. 
\cite{2023XiaTransAct} proposed TransAct, a transformer-based realtime user action sequential model that captures users’ short-term preferences from their most recent actions. 
The encoded user action sequence is represented by a concatenation of the embeddings of the user action type, the content of the pin, and the candidate pin. Then, a standard transformer encoder is employed to aggregate all the information in the user action sequence to represent the user’s short-term preference. 
%%%%%%%%%%%%%%%%%%%%%%%%%%%%%%%
\subsubsection{Long-Term and Short-Term Preferences}
% %%%%%%%%%%%%%%%%%%%%%%%%%%%%%%%%%%%%% both short and long term %%%%%%%%%%%%%%%%%%%%%%%%%%%%%%%%%%%%%
Recognizing that users' long-term and short-term preferences should be captured simultaneously to provide a more comprehensive view of users’ interests and support the construction of a more informative user representation, many studies jointly model long-term and short-term interests, combining stable preference patterns with recent user preferences \citep{yu2019adaptive}. 
A first group of works adopted RNN-based models to encode long-term and short-term user interactions.
For example, \cite{Quadrana2017PersonalizingSession} constructed the user representation through a hierarchical RNN architecture (HRNN) to capture both intra- and inter-session dependencies. Specifically, the model consists of two level-based GRU: a session-level GRU and a user-level GRU. The session-level GRU models user interactions within a session and predicts the next item in the session. The user-level GRU models the user activity across sessions and tracks the evolution of the user interests over time. The user-level GRU propagates information to the session-level GRU, allowing the recommendations to be influenced by the the users' past sessions.
\cite{li2018learning} proposed BINN to discriminatively learn session behaviors and preference behaviors from the users’ interactive behaviors over time for next-item recommendation. The Session Behaviors Learning (SBL) can model the short-term session behaviors by revealing the users’ present consumption motivations using Contextual LSTM (CLSTM). The reference Behaviors Learning (PBL) learns long-term historical stable user preferences using bidirectional CLSTM. 
\cite{kumar2019predicting} proposed JODIE, which employs coupled RNNs to learn the embedding trajectories of a user and an item from an ordered sequence of temporal user-item interactions. Each user and item is assigned static and dynamic embeddings. Static embedding remains unchanged over time to represent long-term stationary properties of users and items. Dynamic embeddings change over time and represent time-varying properties of users and items.

A second group of works adopted 
CNN-based models to encode long-term and short-term user interactions. 
For example, \cite{tang2018personalized} proposed Caser, which uses convolutional filters for
capturing both general preferences and sequential patterns to predict what users will interact with
next. General preferences are modeled through user embeddings based on interacted items. Sequential patterns are learned as local features from the image by embedding a sequence of recent item interactions into an image in the time and latent spaces using convolutional filters.

Other works combined RNN and CNN-based models for long-term and short-term sequential user modeling. For example, \cite{you2019hierarchical} proposed Hierarchical Temporal Convolutional Networks (HierTCN), a hierarchical
deep learning architecture for modeling users’ sequential interactions. It consists of two levels of models: The high-level model uses RNN to aggregate users’ evolving long-term interests across sessions, while the low-level model is implemented with Temporal Convolutional Networks (TCN), utilizing both the long-term interests and the short-term interactions within sessions to predict the next interaction. 

Another group of works adopted GNN-based models to to capture both the long- and short-term
user interests. 
For instance,   
MA-GNN \citep{ma2020memory} applies a GNN for modeling item contextual information in the short-term period, while using a shared memory network to capture long-range dependencies between items.
These representations are then adaptively combined through a gating mechanism to form the final user representation.
In the same vein, \cite{chang2021sequential} proposed SURGE to construct user representations from long-term and short-term behavior sequences. The model first reconstructs a user behavior sequence into an item–item interest graph. Related items are then clustered to represent users’ core interests. Then, an attentive GNN aggregates implicit signals from user behaviors into explicit preference signals that more accurately reflect user preferences.
Finally, dynamic graph pooling is applied to extract users’ current activated core interests from noisy historical behaviors.
 
Another category of
works leverages attention mechanisms to model long-term and short-term preferences.   
For example, \cite{yu2019multi} proposed MARank, a multi-order attentive ranking model that decomposes users’ consuming behavior into long-term and short-term motivations. The long-term preference is modeled through user–item similarity. The short-term preference is captured by extending the user embedding with recent items and modeling individual- and union-level item–item dependencies from multiple views to predict what a user prefers to interact with in the near future. Following this direction, SLi-Rec \citep{yu2019adaptive} constructs user model by integrating long-term and short-term preference representations. Long-term preference captures the users’ general historical interactions through user-item interaction signals, whereas short-term preference is learned from recent behavior sequences using a time-aware and content-aware RNN. The final user representation is generated by adaptively fusing these two components through an attention-based mechanism.

The ability of the self-attention mechanism to capture dependencies among items in the interaction sequence has led to the increasing use of transformer-based models to capture both long-term and short-term preferences.
For instance, \cite{hsu2021retagnn} introduced Relational Temporal Attentive Graph Neural Networks (RetaGNN) for next-item recommendation. It first forms target user–item pairs by pairing each user with their interacted items at different time frames. For each pair, it extracts a local graph around the user and the item. The complete session is used to model the user’s long-term preference, while sub-sessions are used to model short-term preference. These local graphs are passed through a relation-attentive GNN to learn user and item representations. Then, sequential self-attention captures the temporal patterns in both long-term and short-term user preferences and learns the sequential correlation between items within the given session for the final user and item embeddings.
\cite{zhang2018next} proposed AttRec, a self-attention-based sequential recommendation model that 
combines self-attention to model user short-term intent, and a metric learning component to model user long-term preference. 
SASRec \citep{kang2018self} also uses a self-attention mechanism to balance short-term intent and long-term preference.  It seeks to capture long-term dependencies from the user’s interaction history while focusing on the most relevant past actions for next-item prediction. 
As an improved version of SASRec, BERT4Rec \citep{sun2019bert4rec} adopts transformer-based sequential user modeling by constructing the user representation through a bidirectional self-attention network that encodes the user behavior sequence to model long-term and short-term preferences.
TiSASRec \citep{li2020TiSASRectime} improves SASRec by incorporating time-interval information between interactions into the self-attention mechanism. In addition to modeling the sequential order of interacted items, it constructs a time-interval relation for each user based on the time gaps between every pair of items in the historical sequence. 
\cite{zhou2023attention} further extended SASRec by proposing Attention Calibration for Transformer-based Sequential Recommendation (AC-TSR). AC-TSR replaces the standard self-attention layer with an Attention Calibration (AC) layer, which consists of two calibrators: a Spatial Calibrator (SPC) and an Adversarial Calibrator (ADC). SPC incorporates spatial information, including item order and distance, into the attention matrix. ADC mitigates the effect of noisy input by redistributing the attention weights based on each item’s contribution to the next-item prediction. 
FISSA \citep{lin2020fissa} follows SASRec to capture users’ dynamic preferences from their latest interacted item sequence through a local representation learning module. 
To capture users’ global preferences, FISSA further proposes a global representation learning module that applies a location-based attention layer to weigh items by considering their relations to the candidate item. 
Then, a gating module balances the local and global representations
by taking the information of the candidate items into account.
Similarly, SDM \citep{lv2019SDM} uses multi-head self-attention to model short-term user behaviors by capturing multiple user interests in a session. Long-term user preference is encoded through attention and dense fully connected networks based on various types of side information. The model particularly uses a gated fusion module to merge users’ short-term and long-term preferences.  
\cite{2023XiaTransAct} proposed a real-time–batch hybrid ranking approach that encodes user action history. The approach combines TransAct, a transformer-based model for capturing users’ real-time recent actions, with PinnerFormer a transformer-based model which provides batch user embeddings learned over a longer time period. This combination enables to model both short-term and long-term user preferences.
\subsubsection{Limitations and our Contribution}
%%%%%%%%%%%%%%%%%%%%%%%%%%%%%%%
Despite their powerful ability for sequential user modeling and recommendation across a wide spectrum of applications, existing approaches remain limited  in our specific educational context. Given a user’s interaction sequence, a typical sequential user modeling and recommendation model aims to predict and recommend the next likely-to-be-preferred items for this user. Our recommendation goal is different in that we aim  to construct a sequential learner model based on the knowledge concepts that a learner 
did not understand (referred to as DNU concepts) when interacting with a learning material. This learner model is then used to recommend other concepts that can help learners understand their DNU concepts, rather than recommending the next DNU concept in the sequence.  
Moreover, many existing sequential user modeling approaches are supervised, which often requires labeled data, model training, and considerable computational resources. Furthermore, many of these models do not sufficiently consider item features and semantic information, as they often represent items mainly through item IDs.  
Additionally, the attention-based approaches rely on complex attention mechanisms to model user interests, which may increase model complexity.

To address these limitations, in this paper, we propose MR-ConceptGCN as a novel unsupervised sequential learner modeling approach for educational settings. MR-ConceptGCN aims to construct a learner representation that supports the recommendation of concepts to help learners understand or master their DNU concepts. Since the model is unsupervised, it does not require labeled data or supervised training, which reduces computational time and costs. 
Moreover, as text information, e.g., lecture slide content, concept name, and Wikipedia articles could contain useful semantic information about items’ features and users’ interests, MR-ConceptGCN adopts SBERT to generate item embeddings. Furthermore, MR-ConceptGCN incorporates semantic information by using multi-relational graph representation learning to capture relations among concepts, including RELATED\_TO and PREREQUISITE\_TO relationships. In addition, instead of relying on complex attention mechanisms, MR-ConceptGCN uses cosine-based semantic similarity between sequentially interacted items to emphasize relevant concepts, capture dependencies between distant interactions, and downplay less relevant items in the learner’s interaction history. 
Although multi-hop item representations provide useful structural and semantic information, they may be insufficient for modeling fine-grained learner preferences when used alone. Therefore, the representations of the items interacted with by each learner are further refined by incorporating temporal information and semantic relatedness from the learner’s sequential interactions. Concretely, 
MR-ConceptGCN integrates multi-relational graph representation learning with learners’ sequential interactions, allowing the learner model to capture both structural-semantic relations among concepts and the temporal evolution of learner preferences. To model this temporal evolution, MR-ConceptGCN adopts both long-term and short-term learner interests, where each component captures a different aspect of the learner’s interaction history.
The long-term component captures stable preference patterns by modeling semantic relatedness and sequential dependencies among all historically interacted items. Specifically, cosine similarity is used to estimate the relatedness between interacted concepts, allowing greater importance to be assigned to semantically related items. In addition, a mask matrix is employed to preserve chronological order, ensuring that future interactions do not influence past ones while past interactions can contribute to later representations. As a result, the learner representation captures long-term preferences over semantically and sequentially related items.
To capture short-term preferences, MR-ConceptGCN introduces a position-based weighting mechanism that assigns greater importance to more recent interactions in the learner’s history. Consequently, the final learner representation incorporates both correlation-based information and position-based information from the learner’s interacted items, enabling the model to represent both long-term and short-term learner interests.
%%%%%%%%%%%%%%%%%%%%%%%%%%%%%%%%%%%%%%%%%%
\section{Methodology}
\label{methodology}
\begin{figure}[!ht]
	\centering
	\includegraphics[width=1.25\textwidth]{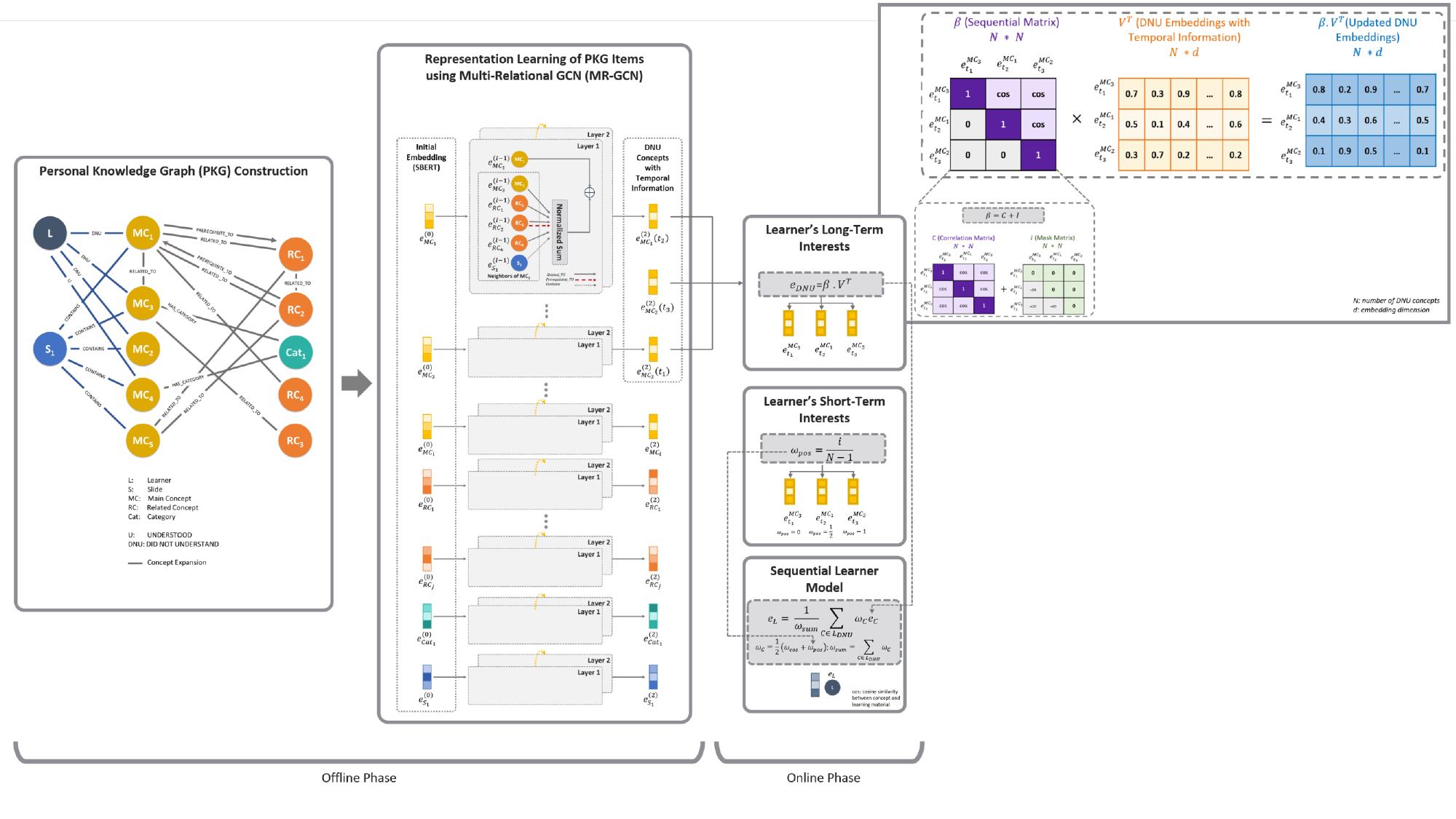}\\
	\caption{Conceptual Architecture of MR-ConceptGCN}
    \label{fig:mr_conceptgcn}
\end{figure}

 In this section, we present the conceptual and technical details of our proposed approach, MR-ConceptGCN. As depicted in Figure \ref{fig:mr_conceptgcn},
 MR-ConceptGCN is divided into two phases, offline and online.
 The offline phase comprises two key steps: (1) PKG construction and (2) representation learning of PKG items using MR-GCN. 
The online phase consists of the sequential learner modeling step based on learner's long-term and short-term interactions.

\subsection{PKG Construction}
\label{Knowledge_graph_construction}
Our PKG contains different nodes representing the following entities: Learner (L), Learning Material (LM), Slide (S), Main Concept (MC), Related Concept (RC), Category (Cat) and edges representing the following relationships: (LM, \textit{CONSISTS\_OF}, S), (S, \textit{CONTAINS}, MC), (MC, \textit{RELATED\_TO}, RC), (MC, \textit{PREREQUISITE\_TO}, MC),  (MC, \textit{PREREQUISITE\_TO}, RC), (MC, \textit{HAS\_CATEGORY}, Cat), (L, \textit{HAS\_READ}, S), (L, \textit{DID NOT UNDERSTAND (DNU) }, MC),  (L, \textit{UNDERSTOOD (U)}, MC). The weight of the edges \textit{CONSISTS\_OF}, \textit{CONTAINS}, \textit{RELATED\_TO}, and \textit{HAS\_CATEGORY} are determined based on the cosine similarity between the two node embeddings using SBERT \citep{reimers2019sentence}. The weight of the edge \textit{PREREQUISITE\_TO} is computed based on a multi-criteria approach that combines the contributions of different features to effectively extract prerequisite relationships, as proposed in \citep{alatrash2025inferring}.

\begin{figure}[!ht]
	\centering
	\includegraphics[width=1.0\textwidth]{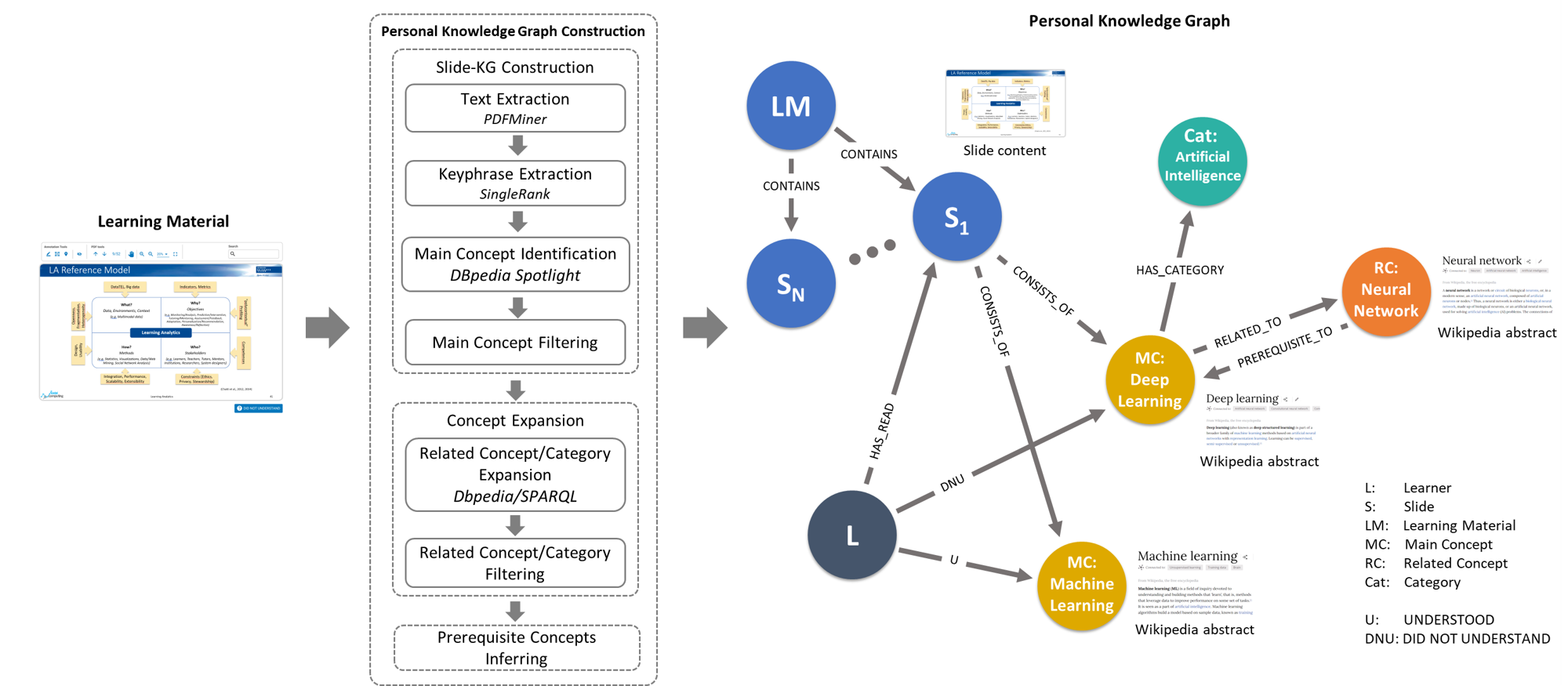}\\
	\caption{Personal Knowledge Graph (PKG) Construction}
	\label{fig:pkg-construction}
\end{figure}

The pipeline for constructing the PKG is illustrated in Figure \ref{fig:pkg-construction}.
Following \citep{ain2025optimized}, the process starts once the teacher uploads the LM, a set of slides, to CourseMapper. Then, the system creates a Slide-KG for each slide, and the combined slide-KGs constitute an LM-KG. 
During this process, the textual content of a slide in the LM is extracted by the python library PDFMiner. Then, the top 15 keyphrases are extracted from each slide using the $SIF Rank_{SqueezeBERT}$ algorithm \citep{ain2023automatic}. 
After that, these keyphrases are linked to specific entities in the DBpedia Spotlight knowledge base \citep{mendes2011dbpedia} to identify the MCs. Next, these MCs are filtered and sorted based on their importance to both Slide and LM. 
This filtering step employs SBERT to generate the initial representation of each entity in the PKG. Then, based on the initial representation, the cosine similarity scores are calculated between the MCs and the slide, as well as between the MCs and the LM, ensuring that the top 5 MCs with the highest importance to the slide and LM are selected. After that, using a locally hosted Wikipedia dump, these MCs undergo an expansion operation to get RCs and Cats based on three steps, namely candidate set creation, candidate set weighting, and candidate pruning. As a result of the concept expansion phase, the top 20 RCs and top 3 Cats for each MC are retained and added to the PKG.
Finally, to infer prerequisites relationships, we follow \citep{alatrash2025inferring} who proposed a multi-criteria unsupervised approach for
automatically inferring concept prerequisites relationships without relying on labeled data.
This approach uses ten criteria based on document-based, Wikipedia hyperlink-based, graph-based, and text-based features, and combines them using a
voting algorithm
to determine the likelihood of one concept being a prerequisite of another one. Using this approach, we infer possible prerequisite relationships among all concept pairs (MC or RC) and update the PKG with all found PREREQUISITE\_TO relationships.

In this work, we address multiple relationship types between concepts, namely \textit{RELATED\_TO} and \textit{PREREQUISITE\_TO}, since we believe that they play a critical role for accurate concept-based learning modeling. Incorporating prerequisite relationships between concepts would ensure that learners have the necessary foundational knowledge before learning advanced concepts.  
After constructing the PKG, we enhance the representations of all PKG items using MR-GCN which considers various relationship types, as discussed in detail in the next section. The enhanced representations of the concepts that a learner did not understand are then utilized by our proposed MR-ConceptGCN model to construct a sequential learner model that combines long-term and short-term learner interactions (see Section \ref{construct_learner_model}).
%%%%%%%%%%%
\subsection{Representation Learning using MR-GCN} \label{MR-GCN}
The goal of this step is to design an MR-GCN model that enhances the representations of PKG items by taking into account various relationships. To this end, we propose and compare two methods: MR-GCN\_RRGCN and MR-GCN\_CompGCN.
\subsubsection{MR-GCN\_RRGCN}
\label{Multi-relational_GCN_method_1}
\begin{figure}[H]
	\centering
	\includegraphics[width=1.4\textwidth]{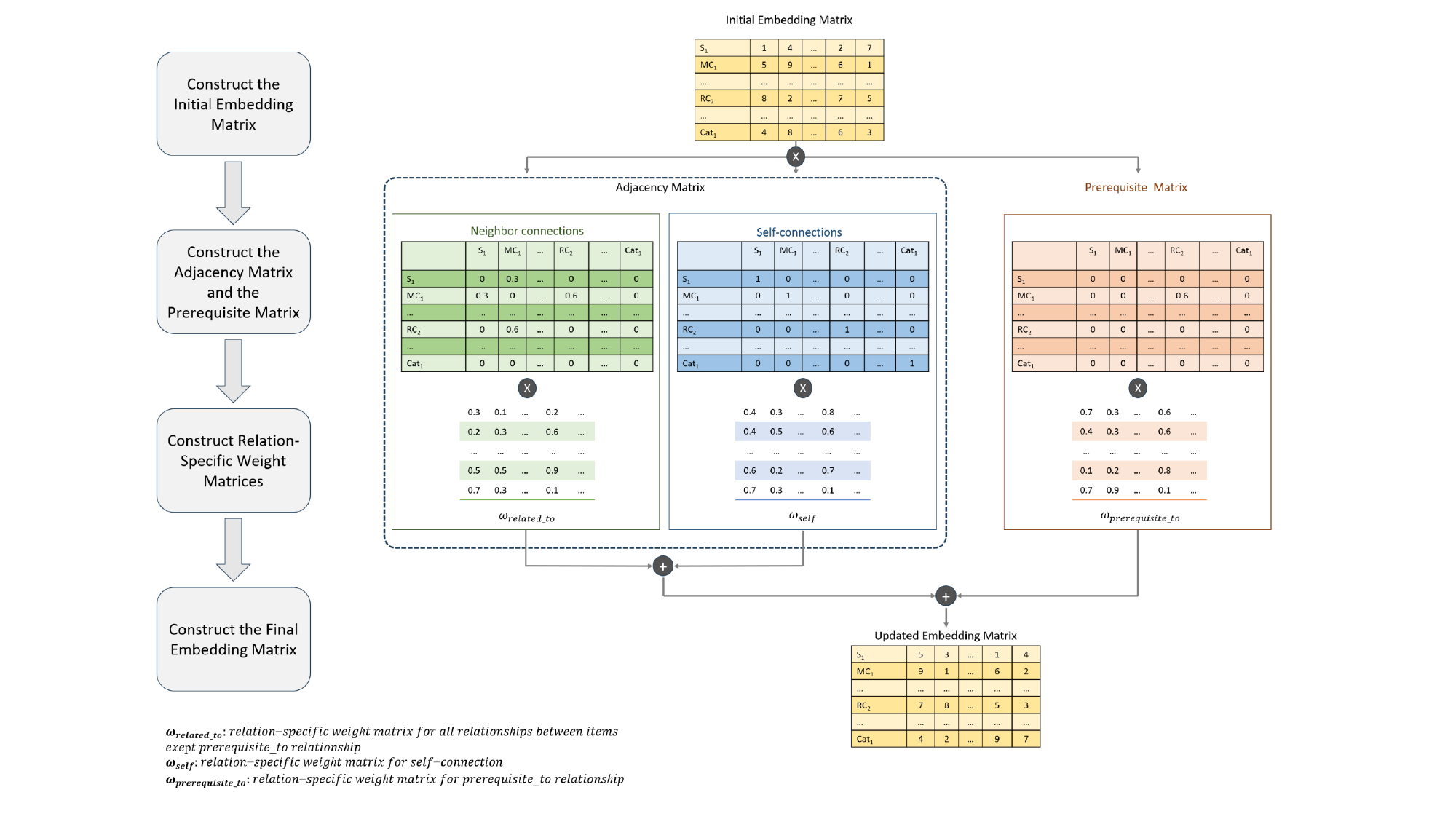}\\
	\caption{Enhanced Representation of PKG Items using MR-GCN\_RRGCN}
	\label{fig:rrgcn}
\end{figure}

We refer to the first method of the MR-GCN model inspired by RR-GCN \citep{degraeve2022rrgcn} as MR-GCN\_RRGCN. In this method, we adapt RR-GCN, which explicitly accounts for multiple relationship types, to enhance the representations of PKG items. Compared to RR-GCN, in MR-GCN\_RRGCN, we use two matrices (i.e., adjacency matrix and prerequisite matrix) and we incorporate
SBERT-based semantic similarity, allowing structural
propagation to be aligned with semantic relatedness between PKG items. 
Enhancing the representation of PKG items in MR-GCN\_RRGCN is mainly achieved through the following steps: (1) construct the initial embedding matrix, (2) construct the adjacency matrix and prerequisite matrix, (3) construct relation-specific weight matrices, and (4) construct the final embedding matrix, as illustrated in Figure \ref{fig:rrgcn}.
%%%%%%%%%%%%
\subsubsubsection{Construct the Initial Embedding Matrix}
\label{Construct_the_initial_embedding_matrix} \leavevmode \\ 
Following \citep{alatrash2024conceptgcn}, the initial embedding matrix for PKG items is constructed using an SBERT-based method \citep{reimers2019sentence}. This method aims to provide initial representations (embeddings) for different PKG items, namely slide (S), main concept (MC), related concept (RC), and category (Cat), that can capture their semantic information, as follows: 
    \begin{equation}
        s = \{s_1, s_2, s_3,..., s_n\}
    \end{equation}
    where $s$ is the set of slide nodes and $n$ is the number of slides belonging to the LM. The embedding of each slide $s_n$ can be represented by applying SBERT on the slide textual content:
    \begin{equation}
        e_{s_n} = SBERT\{slide\_content_n\}
    \end{equation} 
  Then, the set of main concepts ($mc_i$) identified from a slide $s_n$ as well as their expanded related concepts ($rc_j$) and categories ($cat_k$) are represented as follows:
  \begin{equation}
      mc = \{mc_1, mc_2, ... ,mc_i\}
  \end{equation}
   \begin{equation}
      rc = \{rc_1, rc_2, rc_3, ..., rc_j\}
  \end{equation}
    \begin{equation}
      cat = \{cat_1, cat_2, ..., cat_k\}
  \end{equation}
  The embeddings of $mc_i$ and $rc_j$ are derived from the abstracts of their Wikipedia articles and the embedding of $cat_k$ is generated based on the category name using SBERT:
  \begin{equation}
      e_{mc_i} = SBERT \{Wikipedia\_abstract_i\}
  \end{equation}
  \begin{equation}
      e_{rc_j} = SBERT \{Wikipedia\_abstract_j\}
  \end{equation}
  \begin{equation}
      e_{cat_k} = SBERT \{category\_name_k\}
  \end{equation}

\subsubsubsection{Construct the Adjacency Matrix and  the Prerequisite Matrix}
\label{adjacency_matrix_and_connection_matrix}\leavevmode \\ 
We adopt the approach of ConceptGCN \citep{alatrash2024conceptgcn} for constructing the adjacency matrix ($ADJ$) of all PKG items that are connected through 
different relationships except the \textit{PREREQUISITE\_TO} relationship that will be calculated separately in the prerequisite matrix ($PRE$).
The adjacency matrix is represented by the neighbor connection and the self-connection sub-matrices.  The varying degrees of influence exerted by direct neighbors are captured through a simple score function, which can be viewed as an “attention mechanism”. This function combines the symmetric square-root normalization introduced in LightGCN \citep{he2020lightgcn} with an SBERT-based semantic similarity measure. The resulting score reflects the importance of the relationship between two nodes: higher similarity indicates stronger influence from the neighbor, while lower similarity indicates weaker influence. For a directly connected pair of nodes $u$ and $v$ in the PKG, their importance $ADJ_{u,v}$ in the adjacency matrix is calculated as follows:
\begin{equation} \label{eq:adj_weight}
    ADJ_{u,v} = 
     \begin{cases}
    1 & \text{if } u=v \\
    \frac{cos(e_u, e_v)}{\sqrt{|N_u||N_v|}} & \text{if }v \in N_u \\
    0 & \text{otherwise }
    \end{cases}
\end{equation}
where $e_u$ represents the embedding of node $u$, $e_v$ represents the embedding of node $v$, $cos$ is the cosine similarity between these two embedding vectors, $|N_u|$ and $|N_v|$ represent the degree of nodes $u$ and $v$, respectively, that is, the number of directly connected neighbor nodes to $u$ and $v$.
If nodes $u$ and $v$ are not directly connected, the corresponding value is 0. 

We propose a similar approach to construct a prerequisite matrix $PRE$ for \textit{PREREQUISITE\_TO} relationships between a pair of nodes $u$ and $v$ in the PKG. It is important to note that prerequisite relationships are asymmetric and exclude self-connections. If $u$ and $v$ are both concept nodes (MC or RC) and $u$ is a prerequisite of $v$, then the corresponding entry $PRE[u][v]$ in the prerequisite matrix  is
$S^{Pre}_{u,v}$; otherwise, the value is 0. Due to asymmetry, the reverse entry $PRE[v][u]$ is always 0. Similarly, because a concept cannot be a
prerequisite of itself, $S^{Pre}_{u,u}$ is also set to 0. 
The prerequisite matrix is calculated as follows:
\begin{equation} \label{eq:pre_matrix}
    PRE_{u,v} =
    \begin{cases}
    S^{Pre}_{u,v} & \text{if }v \text{ is a prerequisite of }u \\
    0, & \text{otherwise }
    \end{cases}
\end{equation}
where $S^{Pre}_{u,v}$ is a score representing the likelihood that node $u$ is a prerequisite of node $v$, as proposed in \citep{alatrash2025inferring}, with values ranging between 0 and 1. A value closer to 1 indicates a higher likelihood that $u$ is a prerequisite of $v$, while a value closer to 0 indicates a lower likelihood. 
\subsubsubsection{Construct Relation-Specific Weight Matrices}
\label{Construct_relation-specific_weight_matrices}\leavevmode \\
Inspired by RR-GCN \citep{degraeve2022rrgcn}, we generate random relation-specific weight matrices using the Glorot method \citep{glorot2010understanding}, to capture the distinct roles of different relationship types between PKG items. Glorot uniform initialization method is a method for initializing the weights of a neural network to avoid gradient explosion or gradient disappearance during training.
This method is applied at each layer to support effective learning of higher-level feature representations.
For the semantic and structural relationships \textit{CONTAINS}, \textit{CONSISTS\_OF}, \textit{HAS\_CATEGORY}, and \textit{RELATED\_TO}, we generate one random relation-specific weight matrix $W_{related\_to}$. This is because these relationships share a similar mechanism for calculating weights between nodes, and our objective is to differentiate between distinct types of relationships among concepts, namely \textit{RELATED\_TO} and \textit{PREREQUISITE\_TO}.
We model the \textit{PREREQUISITE\_TO} relationship separately due to its directional and asymmetric nature. For this relation, we generate a dedicated random relation-specific weight matrix $W_{prerequisite\_to}$.  Additionally, we include a self-loop weight matrix $W_{self}$, to preserve each node’s inherent features during message passing. To avoid parameter updates during training, all relation-specific weight matrices are fixed after Glorot initialization. This method greatly reduces computational overhead, does not require backpropagation to train the network, and avoids the problem of overfitting the model on small-scale data sets \citep{degraeve2022rrgcn}. The relation-specific weight matrices are constructed as follows for each layer \(l=\{1,2\}\):
        \begin{equation} \label{eq:self_weight_layer_1}
            W^{l}_{self} = Glorot\_uniform(weight\_size,weight\_size)
        \end{equation}
        \begin{equation} \label{eq:ordinary_weight_layer_1}
            W^{l}_{related\_to} = Glorot\_uniform(weight\_size,weight\_size)
        \end{equation}
        \begin{equation} \label{eq:prerequisite_weight_layer_1}
            W^{l}_{prerequisite\_to} = Glorot\_uniform(weight\_size,weight\_size)
        \end{equation}
where \textit{weight\_size} is the size of the node embedding vector and, in our case, we consider a two-layer receptive field. 

\subsubsubsection{Construct the Final Embedding Matrix}\leavevmode \\
In MR-GCN\_RRGCN, we use the embedding of the last layer as the final embedding of PKG items, and in our case, the last layer will be the second layer.
\begin{figure}[!ht]
    \centering
    \begin{minipage}{0.48\textwidth}
        \centering
        \includegraphics[width=6cm]{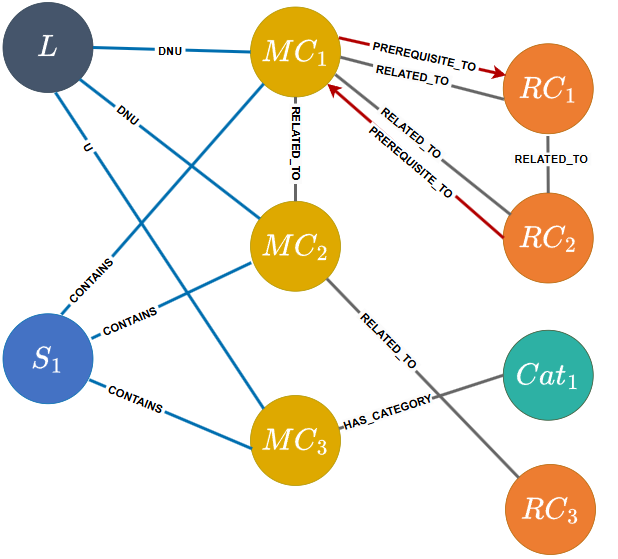}
        \\[1ex]
        \small (a) Sample PKG Structure
    \end{minipage}
    \hfill
    \begin{minipage}{0.48\textwidth}
        \centering
        \includegraphics[width=7cm]{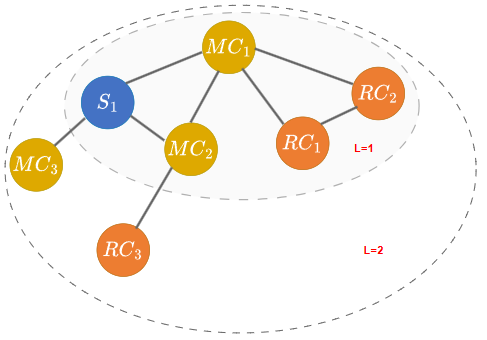}
        \\[1ex]
        \small (b) A Two-Layer Receptive Field Example of Item $MC_1$ in the PKG
    \end{minipage}
    \caption{Sample PKG and a Two-Layer Receptive Field used to Enhance the Embedding of the Target Node $MC_1$}
    \label{fig:sample_pkg_structure_with_two_layer_receptive_field}
\end{figure}
Figure \ref{fig:sample_pkg_structure_with_two_layer_receptive_field} shows an illustrative example of a two-layer receptive field for the target node $MC_{1}$. 
In the left subgraph of this example, we can see that the slide ($S_1$) \textit{CONTAINS} three main concepts ($MC_{1}$, $MC_{2}$, and $MC_{3}$), which are associated with three related concepts ($RC_{1}$, $RC_{2}$, and $RC_{3}$) through \textit{RELATED\_TO} relationships, and has one category ($Cat_{1}$). Also, $MC_{1}$ is a \textit{PREREQUISITE\_TO} $RC_{1}$, and $RC_{2}$ is a \textit{PREREQUISITE\_TO} $MC_{1}$. Learner ($L$) did not understand (\textit{DNU}) $MC_{1}$ and $MC_{2}$, but understood (\textit{U}) $MC_{3}$. The right subgraph of this example shows a two-layer receptive field for the target node $MC_{1}$,  
which describes the path between $MC_{1}$ and its neighbors within two hops by considering different relationship types in the PKG. This high-order connectivity can capture deep semantic information and help generate a richer representation of $MC_{1}$ through propagation and message passing.

Specifically, in the $(l+1)^{th}$ layer, the representation of the node $u$ is updated by aggregating information from its neighbors through considering different relationship types. This aggregation combines the weighted embedding information from the node neighbors in the $l^{th}$ layer and the representation of the node itself (i.e., self-connection). The final node representation is calculated as follows:
\begin{equation}\label{eq:method_1}
\begin{aligned}
e_u^{(l+1)} =\;& W_{self}^{(l+1)} (ADJ_{u,u}e_u^l + PRE_{u,u}e_u^l) 
+ W_{related\_to}^{(l+1)}\sum_{v\in N_{u(related\_to)}} ADJ_{u,v}e_v^l \\
&+ W_{prerequisiste\_to}^{(l+1)}\sum_{v\in N_{u(prerequisiste\_to)}} PRE_{u,v}e_v^l
\end{aligned}
\end{equation}

This equation consists of three parts:
\begin{itemize}
    \item Self-connection part:
        \begin{equation*} \label{eq:self_part_method_1}
            W_{self}^{(l+1)} (ADJ_{u,u}e_u^l + PRE_{u,u}e_u^l)
        \end{equation*}
    where $W_{self}^{l+1}$ is randomly generated by the Glorot method. For the \textit{RELATED\_TO} relationship, $ADJ_{u,u} = 1$ by definition of the self-connection relationship. 
    For the \textit{PREREQUISITE\_TO} relationship, a concept cannot be a prerequisite of itself, therefore, $PRE_{u,u} = 0$. 
    \item Relationships between PKG items, excluding \textit{PREREQUISITE\_TO} relationships:
        \begin{equation*} \label{eq:ordinary_part_method_1}
            W_{related\_to}^{l+1}\sum_{v\in N_{u(related\_to)}}ADJ_{u,v}e_v^l
        \end{equation*}
        where $W_{related\_to}^{l+1}$ represents the relation-specific weight matrix for relationships between PKG items, except \textit{PREREQUISITE\_TO} relationships. $ADJ_{u,v}$ represents the strength of the relationship between node $u$ and node $v$, $e_v^l$ represents the embedding of node $v$ in previous layer, and $N_{u(related\_to)}$ represents the set of all neighbor nodes connected to $u$ through various relationships, except the \textit{PREREQUISITE\_TO} relationship.
     \item \textit{PREREQUISITE\_TO} relationships between concept items in the PKG:
        \begin{equation*} \label{eq:prerequisite_part_method_1}
            W_{prerequisiste\_to}^{l+1} \sum_{v\in N_{u(prerequisiste\_to)}}PRE_{u,v}e_v^l
        \end{equation*}
        where $W_{prerequisiste\_to}^{l+1}$ represents the relation-specific weight matrix for the \textit{PREREQUISITE\_TO} relationship. $PRE_{u,v}$ is used to quantify the likelihood of the existence of \textit{PREREQUISITE\_TO} relationship between concept $u$ and concept $v$, $e_v^l$ represents the embedding of concept $v$ in the previous layer, and $N_{u(prerequisiste\_to)}$ represents the set of all neighbors connected to $u$ through \textit{PREREQUISITE\_TO} relationships.
\end{itemize}
Considering that in self-connection part $ADJ_{u,u} = 1$ and $PRE_{u,u} = 0$, the equation can be rewritten as follows:
\begin{equation} \label{eq:final_method_1}
    e_u^{(l+1)}=W_{self}^{(l+1)} e_u^l  + W_{related\_to}^{(l+1)}\sum_{v\in N_{u(related\_to)}}ADJ_{u,v}e_v^l+W_{prerequisiste\_to}^{(l+1)}\sum_{v\in N_{u(prerequisiste\_to)}}PRE_{u,v}e_v^l
\end{equation}
It is worth noting that, in each layer \(l=\{1,2\}\), we use the three relation-specific weight matrices of the corresponding layer as mentioned earlier in Section \ref{Construct_relation-specific_weight_matrices}.
In our example, the final embedding of $MC_2$ at layer 2 is calculated as follows:
\begin{align*} \label{eq:final_method_ex}
    e_{MC_1}^{2}=W_{self}^2 e_{MC_1}^{1}  + W_{related\_to}^2 (ADJ_{{MC_1},{S_1}}e_{S_1}^1 + ADJ_{{MC_1},{MC_2}}e_{MC_2}^1 + ADJ_{{MC_1},{RC_1}}e_{RC_1}^1 + ADJ_{{MC_1},{RC_2}}e_{RC_2}^1) \\ + W_{prerequisiste\_to}^2 (PRE_{{MC_1},{RC_2}}e_{RC_2}^1)
\end{align*}
%%%%%%%%%%%%%%%%%%%%%%%%%%%%%%%%%%%%%%%%%%%%%%%%%%%%%%%%%
\subsubsection{MR-GCN\_CompGCN}
\label{Multi-relational_GCN-method_2}
\begin{figure}[H]
	\centering
	\includegraphics[width=1.3\textwidth]{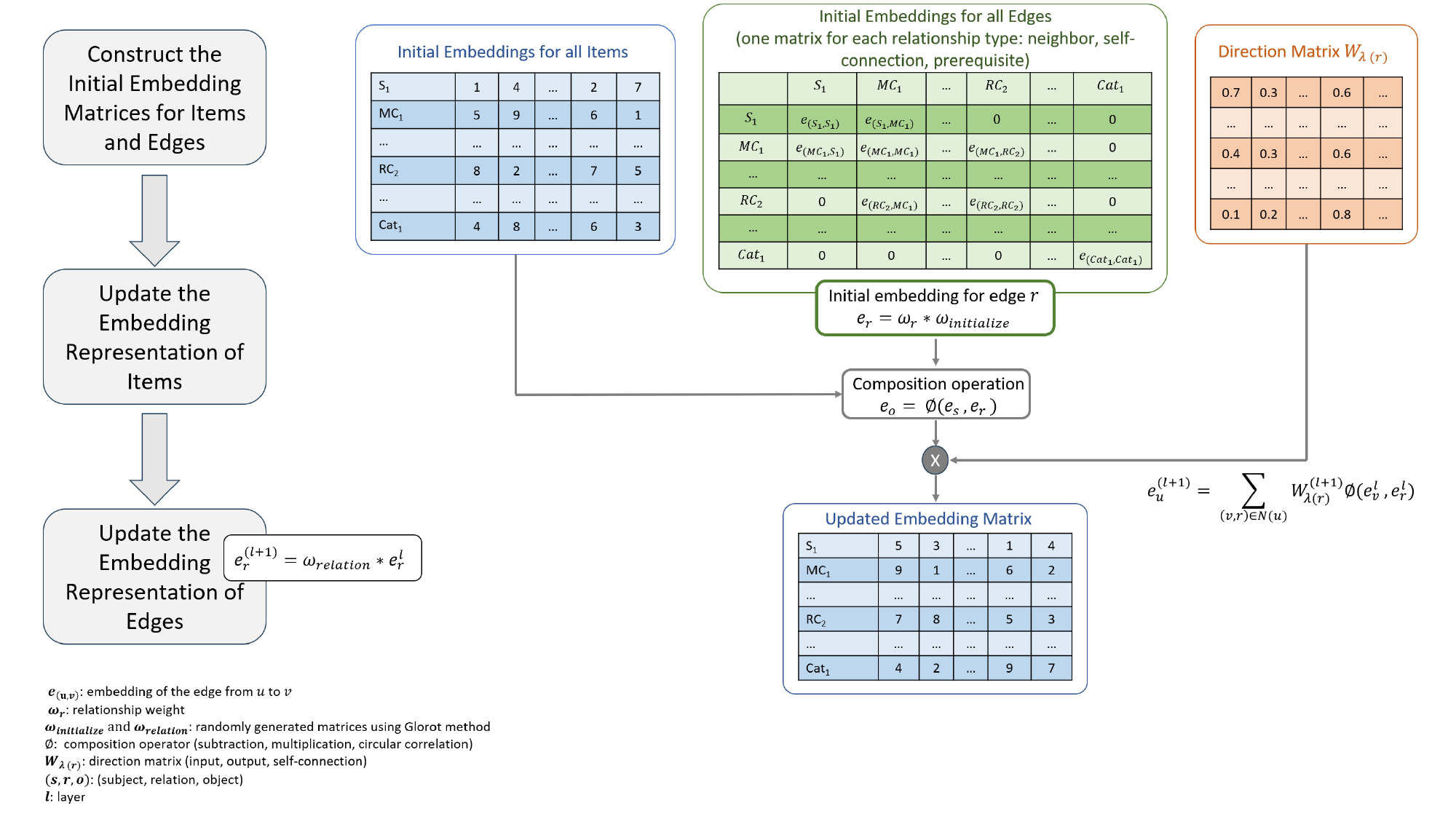}\\
	\caption{Enhanced Representation of PKG Items using MR-GCN\_CompGCN}
	\label{fig:compgcn}
\end{figure}

\begin{figure}[!ht]
	\centering
	\includegraphics[width=0.4\textwidth]{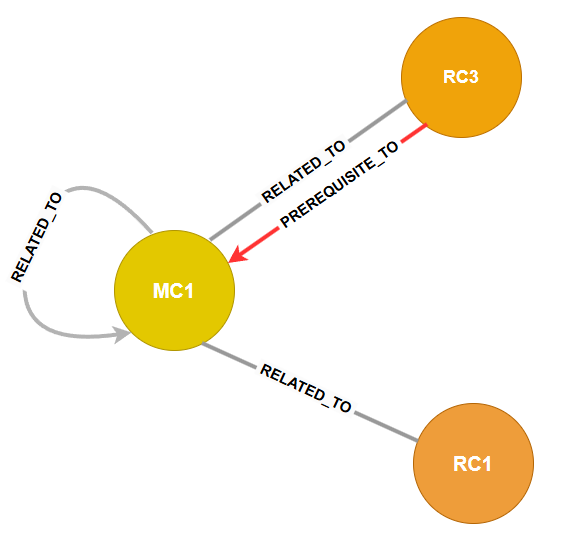}\\
	\caption{Example MR-GCN\_CompGCN }
	\label{fig:example_compgcn}
\end{figure}

The MR-GCN\_RRGCN method discussed above relies on relation-specific weight matrices that are randomly generated for each relationship type at each layer, which might lead to increased model complexity and computation time. In our work, we only differentiate between two types of relationships between concepts, along with self-connections. Therefore, the number of relation-specific weight matrices in our model is relatively small. However, if we introduce more relation types into our model in future work, the number of relation-specific weight matrices will increase. In this case,   
the number of parameters will be large, which might lead to over-parameterization problems. 
To overcome this issue, CompGCN \citep{vashishth2019composition_Compgcn} limits the number of relation-specific weight matrices by generating three relation-specific matrices, namely the Input relation matrix $W_I$, the Output relation matrix $W_O$, and the Self-connection relation matrix $W_S$, and updates them during training. As our approach is fully unsupervised, we converted CompGCN into an unsupervised method by freezing relation representations while retaining node–relation composition.
To this end, we discuss next MR-GCN\_CompGCN, our second proposed method of the MR-GCN model.
In MR-GCN\_CompGCN, embeddings are enriched for both nodes and edges to better capture the semantic information within the node representations. 
Moreover, we incorporate SBERT-based semantic similarity to modulate message passing, allowing structural propagation to be
aligned with semantic relatedness between PKG items.
MR-GCN\_CompGCN is mainly achieved through the following steps: (1) construct the initial embedding matrices for PKG items and edges, (2) update the embedding representation of each item in the PKG, and (3) update the embedding representation of each edge in the PKG,
as depicted in Figure \ref{fig:compgcn}.
\subsubsubsection{Construct the Initial Embedding Matrices for Items and Edges}
\label{Initialize_embedding_edge}\leavevmode \\
Firstly, we compute the initial embeddings for all PKG items (S, MC, RC, and Cat) using SBERT \citep{reimers2019sentence}, following the procedure described in Section~\ref{Construct_the_initial_embedding_matrix}.
Secondly, we compute the initial embeddings for all edges between PKG items. However, since CompGCN requires training and is therefore not directly applicable to our unlabeled setting, we adapt CompGCN by removing the trainable components and reformulating it into an unsupervised approach. Concretely, we propose a way to generate an initial embedding for each edge $e_r$ that leverages the weights of different relationships $w_r$ between PKG items, derived from the adjacency matrix ${ADJ}$ and prerequisite matrix ${PRE}$ introduced in Section ~\ref{adjacency_matrix_and_connection_matrix}. These relationship weights are multiplied by a transformation matrix $w_{initialize}$ randomly initialized using the Glorot method \citep{glorot2010understanding} to obtain customized initial embeddings for each edge $e_r$, as follows:
\begin{equation} \label{eq:edge_embeeding_initialize}
    e_r = w_r*w_{initialize}
\end{equation}
where $w_{initialize}$ is randomly generated using Glorot method as given in equation \ref{eq:edge_initialize_weight}, \textit{weight\_size} is is the size of the node embedding vector, $w_r$ is the relationship weight. In case of different relationship types except \textit{PREREQUISITE\_TO}, the corresponding weight value is obtained from the adjacency matrix ${ADJ}$, otherwise from the prerequisite matrix ${PRE}$, as given in equation \ref{eq:wr}.
\begin{equation} \label{eq:edge_initialize_weight}
    w_{initialize} = Glorot\_uniform(1,weight\_size)
\end{equation}
\begin{equation} \label{eq:wr}
    w_r =
    \begin{cases}
    PRE_r, & \text{if } r \text{ is a PREREQUISITE\_TO relationship} \\
    ADJ_r, & \text{otherwise}
    \end{cases}
\end{equation}
After obtaining the initial embedding matrices for all PKG items and edges, the next step is to use these embeddings to update the representations of the PKG items.
\subsubsubsection{Update the Embedding Representation of Items}\leavevmode \\
To update the embeddings of PKG items, we randomly generate three weight matrices $W_I$, $W_O$, and $W_S$ at each layer $l > 0$, using the Glorot method \citep{glorot2010understanding}. These matrices, collectively referred to as direction matrices and denoted by $W_\lambda$, are constructed as follows:
\begin{equation} \label{eq:w_i}
            W^{l}_{\lambda} = Glorot\_uniform(weight\_size,weight\_size)
        \end{equation}
where \textit{weight\_size} is the size of the node embedding vector.
The embedding of a PKG item $u$ is then updated as follows:
\begin{equation} \label{eq:method_2}
    e_u^{(l+1)} = \sum_{(v,r)\in N(u)} W_{\lambda(r)}^{(l+1)} \phi (e_v^l,e_r^l)
\end{equation}
where $N(u)$ is the set of neighbors that are connected to $u$ through various relationships (Input, Output, and Self-connection) and r is the relationship edge between $u$ and $v$. At each layer $l > 0$, the direction matrix $W_\lambda$ is defined as follows:   
        \begin{equation}
            W_{\lambda(r)}^l = 
            \begin{cases}
            W_I^l, & \text{if  } r \text{ is an input edge to node } u\\
            W_O^l, & \text{if  } r \text{ is a output edge to node } u\\
            W_S^l,  & \text{if  } r \text{ is a self-connection}
            \end{cases}
        \end{equation}
The entity-relation composition operation is defined as follows:
    \begin{equation} \label{eq:entity_relation_composition_operations}
        e_o = \phi(e_s,e_r)
\end{equation}
where $\phi$ is the composition operator and (s, r, and o) represent (subject, relation, and object) in the knowledge graph. The entity-relation composition operator $\phi$ enables the combination of node and relation embeddings to preserve relational semantics during message passing \citep{bordes2013translating}.
We adopt the multiplication operator \citep{yang2014embedding} because it achieves the best performance on the link prediction task \citep{vashishth2019composition_Compgcn}.   
(equation~\ref{eq:multiplication}). 
    \begin{equation} \label{eq:multiplication}
        \phi(e_v^l,e_r^l) = e_v^l*e_r^l
    \end{equation}
In the following example, we show how to update the embedding representation of the PKG item $MC_1$ according to equation \ref{eq:method_2} at layer 1. As shown in Figure \ref{fig:example_compgcn}, there are three nodes, namely $MC_1$, $RC_1$, and $RC_3$ that are connected through 6 edges. 
It is important to mention that different types of PKG relationships except PREREQUISITE\_TO are bidirectional, which means that these edges can be divided into output direction and input direction. The relationships in this example PKG can be expressed as follows:
\begin{itemize}
    \item $r_1$: $MC_1 \;\xrightarrow{\text{RELATED\_TO}}\; MC_1$
    \item $r_2$: $MC_1 \;\xrightarrow{\text{RELATED\_TO}}\; RC_3$
    \item $r_3$: $MC_1 \;\xleftarrow{\text{RELATED\_TO}}\; RC_3$
    \item $r_4$: $MC_1 \;\xleftarrow{\text{PREREQUISITE\_TO}}\; RC_3$
    \item $r_5$: $MC_1 \;\xrightarrow{\text{RELATED\_TO}}\; RC_1$
    \item $r_6$: $MC_1 \;\xleftarrow{\text{RELATED\_TO}}\; RC_1$
\end{itemize}
Therefore, the set of neighbors that are connected to $MC_1$ is:

$N_{(MC_1)} =[(MC_1,r_1),(RC_3,r_2),(RC_3,r_3),(RC_3,r_4),(RC_1,r_5),(RC_1,r_6)]$. The updated embeddings of $MC_1$ at layer 1 can be achieved as follows:

\begin{equation*} \label{eq:example_method_2} \begin{split}
    e_{MC_1}^{1} = &W_S^1 \phi (e_{MC_1}^0,e_{r_1}^0)+W_O^1 \phi (e_{RC_3}^0,e_{r_2}^0) + W_I^1 \phi (e_{RC_3}^0,e_{r_3}^0)+\\&W_I^1 \phi (e_{RC_3}^0,e_{r_4}^0) + W_O^1 \phi (e_{RC_1}^0,e_{r_5}^0) + W_I^1 \phi (e_{RC_1}^0,e_{r_6}^0)
\end{split}
\end{equation*}
where $e_{MC_i}^0$ and $e_{RC_j}^0$ are the initial embeddings for each node, and $e_{r_k}^0$ is the initial embedding for each edge, as computed in Section~\ref{Initialize_embedding_edge}. 
\subsubsubsection{Update the Embedding Representation of Edges}\leavevmode \\
Prior to propagating to the next layer, edge representations are updated to ensure their effectiveness in subsequent node updates. Similar to CompGCN, each edge embedding is multiplied by a relation weight matrix $W_{relation}$. In contrast to CompGCN, where $W_{relation}$ is a learnable parameter, in our model it is randomly initialized using the Glorot method \citep{glorot2010understanding} as given in equation \ref{eq:w_relation}, and kept fixed during propagation. This design choice enables applicability in unlabeled settings and alleviates the challenges of parameter optimization. The edge update is formally defined as:
\begin{equation} \label{eq:edge_update} 
    e_r^{(l+1)} = W_{relation} e_r^l
\end{equation}
\begin{equation} \label{eq:w_relation}
            W_{relation} = Glorot\_uniform(weight\_size,weight\_size)
        \end{equation}
where \textit{weight\_size} is the size of the node embedding vector.
Here, $W_{relation}$ is not generated separately for each relationship type (i.e., Input, Output, Self-connection) or for each layer; instead, a single $W_{relation}$ is shared across all edges and layers, and is multiplied with the corresponding edge embeddings from the previous layer.
%%%%%%%%%%%%%%%%%%%%%%%%%%
\subsection{Learner Modeling based on Sequential Learner Interactions}
\label{construct_learner_model}
Following ConceptGCN \citep{alatrash2024conceptgcn}, we adopt the idea of constructing a learner model based on the concepts that a learner understood (U) or
did not understand (DNU) and extend it by capturing multiple relationships between concepts and integrating
temporal information from learner
interactions with these concepts. In ConceptGCN, the learner model is primarily constructed based on U and DNU concepts. 
Given that concepts already understood by the learner are assigned a value of zero, the learner model can be constructed exclusively from DNU concepts. 
Accordingly, the learner model is represented as a weighted average of the learner’s DNU concepts, where each weight is computed as the cosine similarity between the corresponding concept embedding and the learning material embedding as follows:
  \begin{equation} \label{equation14}
    e_L = \left[\frac{1}{\omega_{sum}}\sum_{c \in DNU} \omega_{c} \: e_c\right]; \\
    \:\omega_{c} = {cos(e_c,e_{lm})}; \:\omega_{sum} = \sum_{c \in DNU} \omega_{c}
\end{equation} 
where $e_c$ is the embedding of concept $c$ and $\omega_{c}$ is its weight in the learning material $lm$.
However, the learner model in ConceptGCN is limited in that it does not capture multiple relationships between concepts. Moreover, it fails to incorporate both long-term and short-term learner interactions with the concepts.
We argue that temporal information from learner interactions is crucial for capturing sequential learning behaviors and for more accurate dynamic learner modeling. Temporal information is the timestamp in which the learner interacted with the concepts.
In practice, learners’ interests and needs evolve dynamically over time. Static interest modeling is therefore insufficient to capture these changes. By incorporating temporal information, the learner model can distinguish between long-term interests (historical interactions) and short-term interests (recent interactions), thereby enabling more accurate learner modeling \citep{hsu2021retagnn,yu2019adaptive}. 
Although our proposed MR-GCN methods (see Section \ref{MR-GCN}) are sufficient to produce final embeddings, multi-hop concept representations alone may be insufficient to capture learners’ fine-grained interests, which can lead to less accurate learner models. Therefore, we further refine the representations of DNU concepts obtained from the preceding offline phase using our MR-GCN model by incorporating sequential interaction signals in the online phase. 
Accordingly, in our proposed approach MR-ConceptGCN, the learner model embedding is generated in the online phase by leveraging individual learners' long and short-term interactions to further enhance the embeddings of the DNU concepts that build the base of the learner model. In this phase, learners interact with learning materials in CourseMapper by explicitly marking concepts as U or DNU. Based on this feedback, a learner model ($L$) is constructed to capture the learner’s knowledge state, derived from interacted U and DNU concepts. After that, the learner model embedding ($e_L$) is generated based on the sequential order of DNU concepts. 
This allows to integrate time information from learner interactions and to capture both learner's long-term and short-term interests. 
The long-term interests are calculated based on the semantic correlation and the sequential order between DNU concepts in a given learner-concept interaction sequence. The short-term interactions are the DNU concepts most recently marked by a learner, which are more representative of their current interests.
To achieve this, our learner modeling approach involves three steps: (1) construct the learner model based on learner's historical interactions with the concepts, (2) calculate the learner's long-term interests and update the DNU concept embeddings based on semantic and sequential information, and (3) calculate the learner's short-term interests and generate the final learner model embedding.
\subsubsection{Construct the Learner Model based on Historical Interactions}
As described earlier, the learner model $L$ is constructed based on the learner's U and DNU concepts. The model is represented as a binary vector, where each concept is assigned a value of 1 if marked as DNU and 0 if marked as U. 
The timestamp information is incorporated into the representation of the learner model. For each interaction in which a learner marks a concept as DNU, a timestamp is recorded to capture the time of interaction. This temporal information allows the model to differentiate between short-term interests (recent interactions) and long-term interests (historical interactions). In the example given below, the learner did not understand concepts $MC_3$, $MC_1$, and $MC_2$, but understood concept $MC_4$.
Thus, the learner model $L$ can be represented as follows:
\begin{equation*} \label{eq:enhanced_learner_model} 
    L = [(1,t_1)_{C_3},(1,t_2)_{C_1},(0,-)_{C_4},(1,t_3)_{C_2}]
\end{equation*}
Here, $t_1$, $t_2$, and $t_3$ denote the timestamps at which the learner marked $MC_3$, $MC_1$, and $MC_2$ as DNU, respectively, while $MC_4$ (marked as U) does not have an associated timestamp. Incorporating this temporal information provides a clearer view of the learner’s evolving needs. The learner model can also be represented only by DNU concepts, which can be expressed as follows:
\begin{equation} \label{eq:enhanced_DNU_learner_model} 
    L_{DNU} = [(1,t_1)_{C_3},(1,t_2)_{C_1},(1,t_3)_{C_2}]
\end{equation}
\subsubsection{Calculate the Learner's Long-Term Interests} \label{longterm}
Our goal is to capture learners’ long-term interests based on DNU concepts in a given learner-concept interaction sequence. 
To dynamically update the learner model and capture long-term interactions, we employ a sequential matrix. The sequential matrix integrates a correlation matrix and a mask matrix to update the embeddings of DNU concepts. This mechanism captures both the semantic correlation and the sequential order between DNU concepts in the learner model.
The use of a correlation matrix enables the capture of a learner’s topical focus by assigning greater importance to semantically related concepts. This matrix leverages cosine similarity to infer the degree of semantic relatedness among DNU concepts, allowing the model to give more weight to concepts from the same category.  
For example, suppose the learner's list of DNU concepts include "supervised learning", "IoT", and "Bayesian classifier". By calculating cosine similarity, we find that "supervised learning" and "Bayesian classifier" have a high similarity, indicating they belong to the same category machine learning (ML). In contrast, "IoT" has a low similarity to these two concepts, as it belongs to a different category.    
This suggests that the learner may be more interested in the ML category. Accordingly, when constructing the learner model embedding, "supervised learning" and "Bayesian classifier" will receive more importance so that they contribute more to the final learner representation. In this way, the recommendation of new concepts or new learning resources that can help the learner understand the marked DNU concepts would contain items more closely related to "supervised learning" and "Bayesian classifier".

Furthermore, the use of a mask matrix preserves the temporal consistency of learner interactions. This matrix ensures that the sequential order of a learner’s interaction history (marking concepts as DNU) adheres to the natural causal rule: past events can influence the future, but future events cannot influence the past. Inspired by RetaGNN \citep{hsu2021retagnn}, the mask matrix is used to restrict the flow of information in the sequence data, preventing future interactions from affecting past ones, while past interactions affect the future. 
For example, suppose a learner first marks "supervised learning" as DNU, then "IoT", and later "Bayesian classifier". The mask matrix is used to prevent that "supervised learning" is affected by "IoT" and "Bayesian classifier", while still allowing past interactions ("supervised learning" and "IoT") to influence "Bayesian classifier". When updating embeddings of the DNU concepts in the learner model, the mask matrix guarantees that the representation of "supervised learning" contains information from "IoT" and "Bayesian classifier", so that the recommendation of new items that can help the learner understand the marked DNU concepts would contain items more closely related to the more recent concepts "IoT" and "Bayesian classifier".   
By enforcing this temporal constraint, the mask matrix helps to capture the sequential order of interaction events more faithfully, thereby improving its ability to model dynamic learner interactions.

The sequential matrix which integrates a correlation matrix and a mask matrix thus ensures giving higher importance to a recent DNU concept that is semantically related to other DNU concepts in the learner model. In our example, using a sequential matrix to update the embeddings of the chronologically ordered DNU concepts ”supervised learning”, ”IoT”, and ”Bayesian classifier" ensures that the recommendation of new items that can help
the learner understand the DNU concepts would contain items more closely related to ”Bayesian classifier”.   
 
Formally, we update the embeddings of the DNU concepts in $L_{DNU}$ by using a sequential matrix $\beta$, as follows:
\begin{equation} \label{eq:update_dnu_embedding} 
    e_{DNU} = \beta \cdot V^{\mathsf{T}}
\end{equation}
where,
\begin{itemize}
\item $e_{DNU}$ denotes the output matrix, composed of the updated concept embeddings $(e_{C_1}, e_{C_2}, e_{C_3})$. 
    \item $V$ represents the embeddings of the time-ordered DNU concepts obtained using MR-GCN that incorporates multiple relationship
types between concepts. $V$ can be represented as:
        \begin{equation}\label{eq:VL} 
            V=[e^{C_3}_{t1},e^{C_1}_{t2},e^{C_2}_{t3}]
        \end{equation} 
    \item $\beta$ represents the sequential matrix of learner's DNU concepts. It can be calculated as follows:
        \begin{equation}\label{eq:sequential_matrix}
            \beta = C+I
        \end{equation}
        and replace each entry $-\infty$ in $\beta$ with 0. $C$ represents the correlation matrix consisting of cosine similarities $(cos)$ between all pairs of concepts in the DNU list $L_{DNU}$. 
        \begin{equation}\label{eq:correlation_matrix}
C_{ij}=\cos\!\big(
e^{c_x}_{t_i},\;
e^{c_y}_{t_j}
\big)
\end{equation}
where $e^{c_x}_{t_i}$ and $e^{c_y}_{t_j}$ denote the embeddings $e$ of the DNU concepts ${c_x}$ and ${c_y}$ marked by the learner at timestamps $t_i$ and $t_j$, respectively. 

$I$ represents the mask matrix, calculated as follows: 
        \begin{equation} \label{eq:mask_matrix}
            I_{t_it_j} =
            \begin{cases}
            0, & t_i<=t_j \\
            -\infty, & t_i>t_j
            \end{cases}
        \end{equation}
        Referring to the concept embeddings and their timestamps in the list of DNU concepts $V$, we can explain the mask matrix as follows:
        \begin{itemize}
            \item $t_1<t_2<t_3$ time order in which the concepts  concepts $C_3$, $C_1$, and $C_2$ are marked as DNU:
            \item $I_{t_1t_2}$: Can an interaction ($L-DNU->C_3$) that happened at $t_1$ influence an interaction ($L-DNU->C_1$) that happened at $t_2$? The answer is yes, because $t_1<t_2$, which means that interaction ($L-DNU->C_3$) occurs earlier than interaction ($L-DNU->C_1$), so interaction ($L-DNU->C_3$) can affect interaction ($L-DNU->C_1$). The value of $I_{t_1t_2}$ in the mask matrix will be then 0;
            \item $I_{t_2t_1}$: Can an interaction ($L-DNU->C_1$) that happened at $t_2$ influence an interaction ($L-DNU->C_3$) that happened at $t_1$? The answer is no, because $t_2>t_1$, which means that interaction ($L-DNU->C_1$) occurs later than interaction ($L-DNU->C_3$), so interaction ($L-DNU->C_1$) as a future interaction cannot affect the past interaction ($L-DNU->C_3$). The value of $I_{t_2t_1}$ in the mask matrix will be then $-\infty$;
        \end{itemize}
\end{itemize}
The following example illustrates how the embeddings of the DNU concepts $C_3$, $C_1$, and $C_2$ with timestamps $t_1$, $t_2$, and $t_3$, respectively, are updated. 
\begin{enumerate}
    \item Step 1: Take the embeddings of $C_3$, $C_1$, and $C_2$ as calculated in the offline phase using MR-GCN\_CompGCN to form $V$, according to equation \ref{eq:VL}.  
  \item Step 2: Calculate the correlation matrix $C$, according to equation \ref{eq:correlation_matrix}:
        \begin{equation}
            C=\begin{bmatrix}
            1 & cos(e_{t_1}^{C_3},e_{t_2}^{C_1}) & cos(e_{t_1}^{C_3},e_{t_3}^{C_2}) \\
            cos(e_{t_2}^{C_1},e_{t_1}^{C_3}) & 1 & cos(e_{t_2}^{C_1},e_{t_3}^{C_2}) \\
            cos(e_{t_3}^{C_2},e_{t_1}^{C_3}) & cos(e_{t_3}^{C_2},e_{t_2}^{C_1}) & 1
            \end{bmatrix}
        \end{equation}
  \item Step 3: Calculate the mask matrix $I$, according to equation \ref{eq:mask_matrix}:
        \begin{equation}
            I=\begin{bmatrix}
            0 & 0 & 0 \\
            -\infty & 0 & 0 \\
            -\infty & -\infty & 0
            \end{bmatrix}
        \end{equation}
  \item Step 4: Calculate the sequential matrix $\beta$, according to equation \ref{eq:sequential_matrix}:
          \begin{equation}
            \begin{split}
            \beta &= C +I \\
                    &= \begin{bmatrix}
                        1 & cos(e_{t_1}^{C_3},e_{t_2}^{C_1}) & cos(e_{t_1}^{C_3},e_{t_3}^{C_2}) \\
                        cos(e_{t_2}^{C_1},e_{t_1}^{C_3}) & 1 & cos(e_{t_2}^{C_1},e_{t_3}^{C_2}) \\
                        cos(e_{t_3}^{C_2},e_{t_1}^{C_3}) & cos(e_{t_3}^{C_2},e_{t_2}^{C_1}) & 1
                        \end{bmatrix}+
                        \begin{bmatrix}
                        0 & 0 & 0 \\
                        -\infty & 0 & 0 \\
                        -\infty & -\infty & 0
                        \end{bmatrix}\\
                    &=\begin{bmatrix}
                        1 & cos(e_{t_1}^{C_3},e_{t_2}^{C_1}) & cos(e_{t_1}^{C_3},e_{t_3}^{C_2}) \\
                        -\infty & 1 & cos(e_{t_2}^{C_1},e_{t_3}^{C_2}) \\
                        -\infty & -\infty & 1
                        \end{bmatrix}
            \end{split}
        \end{equation}
        and then replace each entry $-\infty$ in $\beta$ with 0:
          \begin{equation}
            \beta =\begin{bmatrix}
                        1 & cos(e_{t_1}^{C_3},e_{t_2}^{C_1}) & cos(e_{t_1}^{C_3},e_{t_3}^{C_2}) \\
                        0 & 1 & cos(e_{t_2}^{C_1},e_{t_3}^{C_2}) \\
                        0 & 0 & 1
                        \end{bmatrix}
        \end{equation}
    \item Step 5: Multiply $\beta$ and $V^T$ to get the updated embeddings of the DNU concepts, according to equation \ref{eq:update_dnu_embedding}.
\end{enumerate}
After the above series of operations, we have successfully captured the learner's long-term interests based on the semantic correlation and the sequential
order among DNU concepts in the learner model. Next, we will calculate the learner's short-term interests and combine them with the learner's long-term interests to generate the final learner model embedding.
\subsubsection{Calculate the Learner's Short-Term Interests and Generate the Learner Model Embedding}
Our approach for generating the learner model embedding $e_L$ is inspired by ConceptGCN \citep{alatrash2024conceptgcn}, which we extend by incorporating temporal information to model long-term and short-term interests. We assume that the DNU concepts most recently marked by a learner are more representative of their current interests. This reflects the intuition that recent DNU concepts are more critical for the learner to understand, and therefore should contribute more strongly to the learner’s representation.
To better capture the learner's short-term interests when generating $e_L$, we introduce a position weight $w_{pos}$, determined by the chronological order of concepts in the DNU list.
The position weight $w_{pos}$ reflects the order in which each DNU concept is interacted with by the learner, giving a higher weight to the most recently marked concepts, thereby better capturing the learner's current interests. The position weight $w_{pos_i}$ is defined as:
\begin{equation} \label{eq:position_weight}
    w_{pos_i} = \frac{i}{N-1}
\end{equation}
where $i$ represents the position index of the DNU concept in the interaction sequence.  $N$ is the total number of concepts marked as DNU by the learner. In the list of DNU concepts $L_{DNU}$ in our example, the position $i$ of $C_3$ "supervised learning" in timestamp $t_1$ is 0 (and $w_{pos_1}$ is 0), the position $i$ of $C_1$ "IoT" in timestamp $t_2$ is 1 (and $w_{pos_2}$ is $\frac{1}{2}$), and the position $i$ of $C_2$ ”Bayesian classifier” in timestamp $t_3$ is 2 (and $w_{pos_3}$ is 1). Accordingly, when constructing the learner
model embedding, ”IoT” and ”Bayesian classifier” will receive more
importance so that they contribute more to the final learner representation. In this
way, the recommendation of new items that can help
the learner understand the marked DNU concepts would contain many items closely
related to ”Bayesian classifier” and also few items related to ”IoT”, although ”IoT” does not belong to the same category as ”Bayesian classifier” and "supervised learning". Thus, taking learner’s short-term interests into account ensures that items from different categories can be recommended. 

Finally, based on the learner's long-term interests, as calculated in Section \ref{longterm}, together with the incorporation of positional weights $w_{pos}$ to capture the learner's short-term interests, we construct the learner model $e_L$ as a weighted average of the learner’s DNU concept embeddings.
Formally, the final embedding of the learner model is defined as:
\begin{equation} \label{eq:learner_model}
    e_L = \frac{1}{w_{sum}}\sum_{C\in L_{DNU}}w_Ce_C; w_C = \frac{1}{2}(w_{cos}+w_{pos}); w_{sum}=\sum_{C\in L_{DNU}}w_C
\end{equation}
where $e_C$ denotes the embedding of DNU concept $C$ after calculating the long-term interest. $w_{cos}$ is the cosine similarity weight between the concept embedding and the learning material embedding. $w_{pos}$ is the position-based weight reflecting recency. $w_{sum}$ is the normalization factor. $w_C$ is the weight of each concept which integrates both semantic relatedness (via $w_{cos}$) and temporal recency (via $w_{pos}$), ensuring that concepts closely related to the learning material and recently marked as DNU receive higher importance.  

In summary, our proposed approach MR-ConceptGCN constructs a sequential learner model that captures both learner's long-term and short-term interests by considering multiple relationship
types between concepts and combining semantic relatedness, sequential information, and positional order.
%%%%%%%%%%%%%%%%%%%%%%%%%%
\section{Evaluation and Results}
\label{online_evaluation}
We conducted an online user study ($n$=31) to collect users’ opinions regarding our proposed approach MR-ConceptGCN for sequential learner modeling. We evaluated three sequential learner models constructed using concept embeddings generated by MR-GCN\_CompGCN, MR-GCN\_RRGCN, and ConceptGCN \citep{alatrash2024conceptgcn}.
Unlike MR-GCN\_CompGCN and MR-GCN\_RRGCN, ConceptGCN does not take into account different relation types (i.e., RELATED\_TO and PREREQUISITE\_TO) when enhancing the concept embeddings.
The evaluation is conducted indirectly by assessing the quality of the recommendations of new concepts that can help learners understand their marked DNU concepts. Since these recommendations are generated based on the constructed learner models, their quality serves as an indicator of how accurately the sequential learner models capture user characteristics. We anticipate that more accurate learner model representations will enhance the relevance and impact of the recommendations. We developed an educational recommender system in CourseMapper to recommend concepts that best align with a user's sequential learner model. To this end, we first compute the cosine similarity between the learner model embedding $e_L$ and the embedding of each candidate concept. A candidate concept may be either a main concept or a related concept, provided that it has not already been marked by the learner as understood (U) or not understood (DNU). The ten concepts with the highest similarity scores are then recommended. Finally, we identify and visualize all existing \textit{PREREQUISITE\_TO} paths among the recommended concepts.
%%%%%%%%%%%%%%%%%%%%%%%%%%
\subsection{Study Design}
The study involved 31 participants from various age groups and academic backgrounds. As summarized in Table~\ref{tab:demography}, 16 participants were female and 15 were male. Only two participants were over 35 years old, while the remaining participants were between 18 and 35. Their educational backgrounds included Bachelor's, Master's, and Ph.D. degrees. Most participants were studying Computer Engineering or Computer Science, and reported good familiarity with recommender systems and machine learning.
The study was conducted online using screen sharing via Microsoft Teams and Zoom, and took approximately one hour for each participant. 
The participants’ answers were recorded, and their consent was obtained to record the session. All collected data were kept confidential and were not shared with third parties. All personally identifiable information was anonymized.

\begin{figure}[h]
    \centering
    \includegraphics[width=1.0\textwidth]{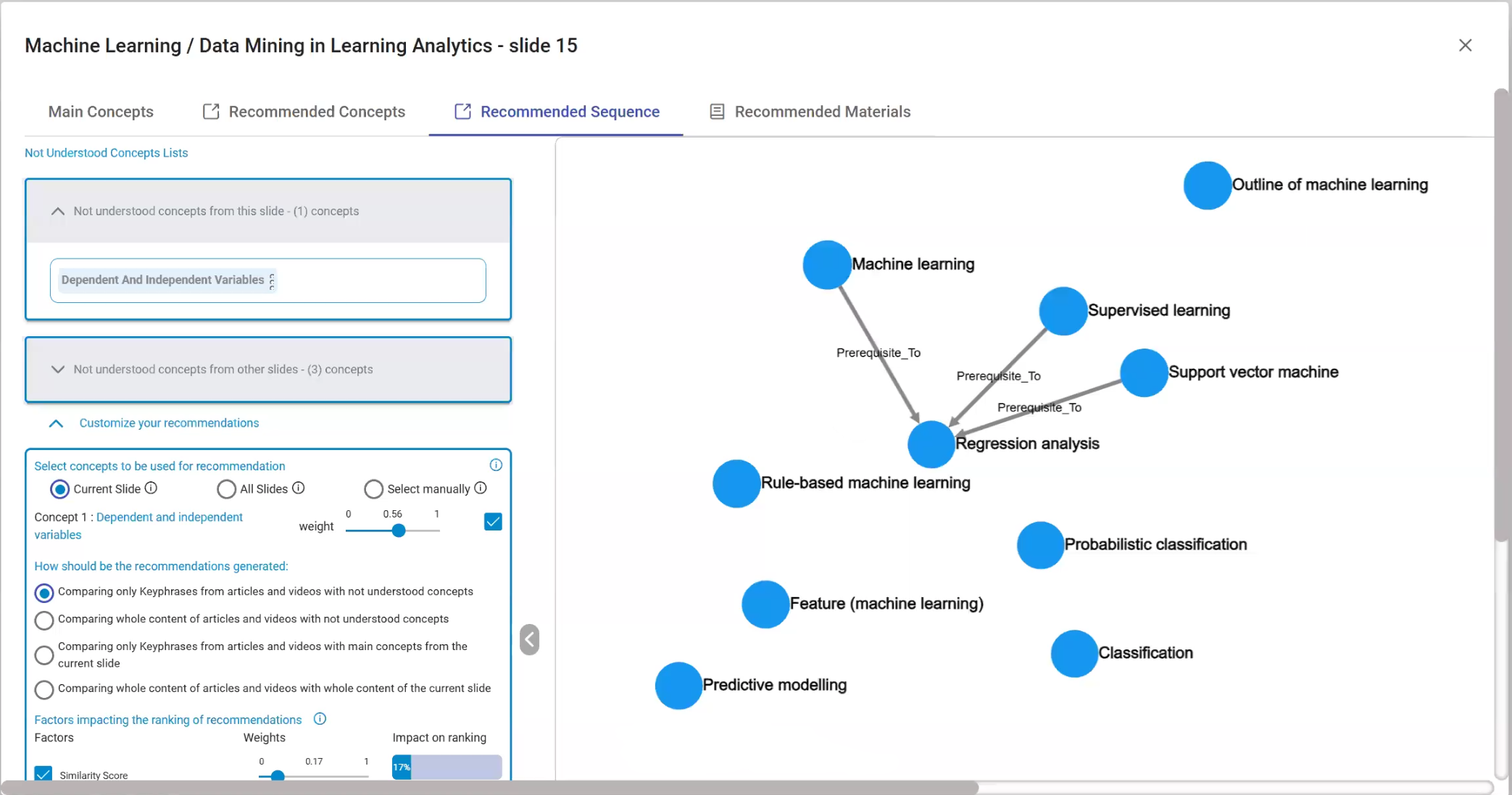}
    \caption{Recommendation of Ten Concepts Including Existing Prerequisite Relationships}
    \label{fig:screenshot}
\end{figure}

In the initial phase, participants were provided with a comprehensive explanation of the purpose and content of the evaluation.  
Subsequently, they were introduced to the system through an interactive video. After confirming their understanding, they were asked to complete their demographic profiles. Participants then performed a task on CourseMapper via remote screen control. The task involved reading a learning material on “Machine Learning / Data Mining in Learning Analytics” sourced from a course offered by our department. The material consisted of a combination of text, images, mathematical formulas, code snippets.
During the task, participants were asked to identify three concepts within the learning material that they did not understand (DNU concepts). After that, the system recommends ten concepts intended to support participants in understanding their marked DNU concepts and visualizes any existing prerequisite relationships among the recommended concepts, as shown in Figure \ref{fig:screenshot}.
Participants were then instructed to review the recommended concepts, read their corresponding Wikipedia descriptions, and complete a questionnaire to assess the quality of the recommended concepts.
This process was repeated for all three learner models, meaning that each participant received recommendations generated by each model. We randomized the order of models to prevent order effects in a within-subject design. 

\begin{table}[htbp]
\centering
\small
\caption{Demographic information of the participants}
\label{tab:demography}
\begin{tabularx}{\textwidth}{@{}l X c@{}}
\toprule
\textbf{Demography} & \textbf{Category} & \textbf{Frequency} \\
\midrule

\textbf{Age} & 18--20 & 1 \\
& 21--25 & 10 \\
& 26--30 & 16 \\
& 31--35 & 2 \\
& Above 35 & 2 \\

\midrule
\textbf{Gender} & Female & 16 \\
& Male & 15 \\

\midrule
\textbf{Education Level} & Bachelor & 5 \\
& Master & 20 \\
& Ph.D. & 6 \\

\midrule
\textbf{Field of Study} & Applied Computer Science & 1 \\
& Automation and Safety & 1 \\
& Business Administration & 1 \\
& Chemistry & 3 \\
& Computer Engineering & 10 \\
& Computer Science & 7 \\
& Data Science & 1 \\
& Embedded Systems Engineering & 2 \\
& Informatik & 1 \\
& Logistics & 1 \\
& Software Engineering & 1 \\
& Teaching Mathematics and Physics & 1 \\
& Telecommunications & 1 \\

\midrule
\textbf{Recommender System Familiarity} & Not at all & 3 \\
& A few & 3 \\
& Average & 5 \\
& Quite & 9 \\
& Very & 11 \\

\midrule
\textbf{Machine Learning Familiarity} & Not at all & 4 \\
& A few & 4 \\
& Average & 6 \\
& Quite & 14 \\
& Very & 3 \\

\bottomrule
\end{tabularx}
\end{table}
The questionnaire was administered using Google Forms. 
We utilized the ResQue evaluation framework \citep{puUsercentricEvaluationFramework2011} to 
provide a comprehensive evaluation of the recommendation performance, which serves as an implicit assessment of the learner model quality. Concretely, we evaluated users’ perceived
benefits in terms of perceived system qualities (perceived accuracy, diversity, information sufficiency), users' beliefs (perceived usefulness), users' attitudes (overall
satisfaction), and behavioral intentions (use intention), on a 5-point Likert scale. One question was designed for each construct, using statements adapted from the ResQue framework.
\begin{itemize}
    \item Perceived Accuracy: The concepts recommended to me matched my interests.
    \item Diversity: The concepts recommended to me are diverse.
    \item Information Sufficiency: The information provided for the recommended items is sufficient for me to understand my "Not Understood" concepts.
    \item Perceived Usefulness: The recommender system gave me good suggestions.
    \item Overall Satisfaction: Overall, I am satisfied with the provided recommendations.
    \item Use Intention: I will use this recommender system frequently.
\end{itemize}
In addition to the items taken from the ResQue framework, for each model we asked participants “How many of the recommended concepts do you feel are relevant?”. The responses to this question were used to calculate the precision (Precision@10) for each model. 
We used the Friedman test, which is appropriate for comparing more than two related conditions when the variable of interest is ordinal. We tested for statistical significance at $\alpha = 0.05$ level.
In the end of the session, we asked the participants two open-ended questions to gather their suggestions for potential improvements. These were: \textbf{(1)} \textit{Which information provided for the recommended items helped you understand your “Not Understood” concepts? And why?} \textbf{(2)} \textit{What additional information do you expect that can help you understand your “Not Understood” concepts?} 
%%%%%%%%%%%%%%%%%%%%%%%%%%%%%%
\subsection{Results and Analysis}
We evaluated the performance of the three sequential learner models constructed using concept embeddings generated with MR-GCN\_CompGCN, MR-GCN\_RRGCN, and ConceptGCN by assessing the potential impact of the recommendations generated based on the three models on system accuracy as well as users’ perceptions
of the recommender system in terms of several important user-centric aspects including perceived accuracy, diversity, information sufficiency, perceived usefulness, overall satisfaction, and use intention. 
To measure system accuracy, we calculated Precision@10 for the three models using participants’ responses to the question “How many of the recommended concepts do you feel are relevant?”. The results indicate that all models achieved high precision scores, implying their provision of relevant recommendations. Specifically, ConceptGCN obtained the highest precision score ($68\%$), followed by MR-GCN\_CompGCN ($64\%$) and MR-GCN\_RRGCN ($63\%$).

\begin{figure}[!htbp]
\centering
\begin{subfigure}{0.45\linewidth}
    \centering
    \includegraphics[width=\linewidth]{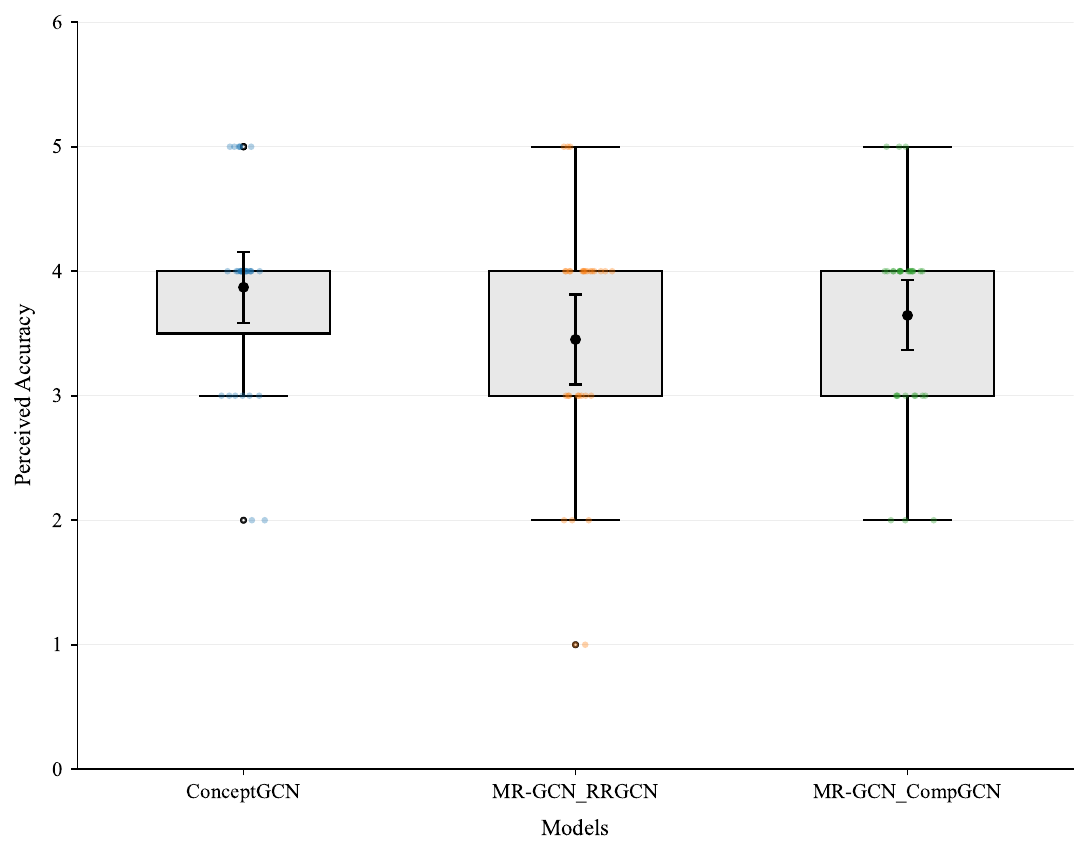}
    \caption*{Perceived Accuracy}
\end{subfigure}
\hfill
\begin{subfigure}{0.45\linewidth}
    \centering
    \includegraphics[width=\linewidth]{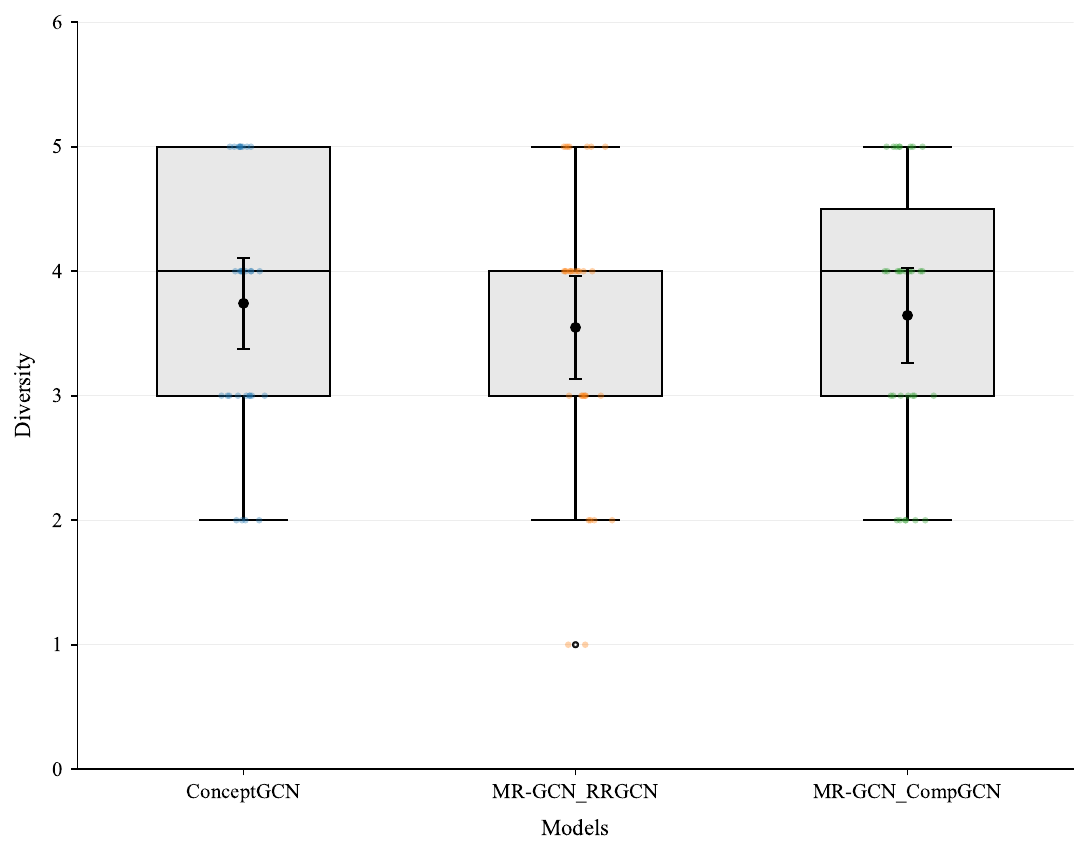}
    \caption*{Diversity}
\end{subfigure}
\vspace{0.15cm}
\begin{subfigure}{0.45\linewidth}
    \centering
    \includegraphics[width=\linewidth]{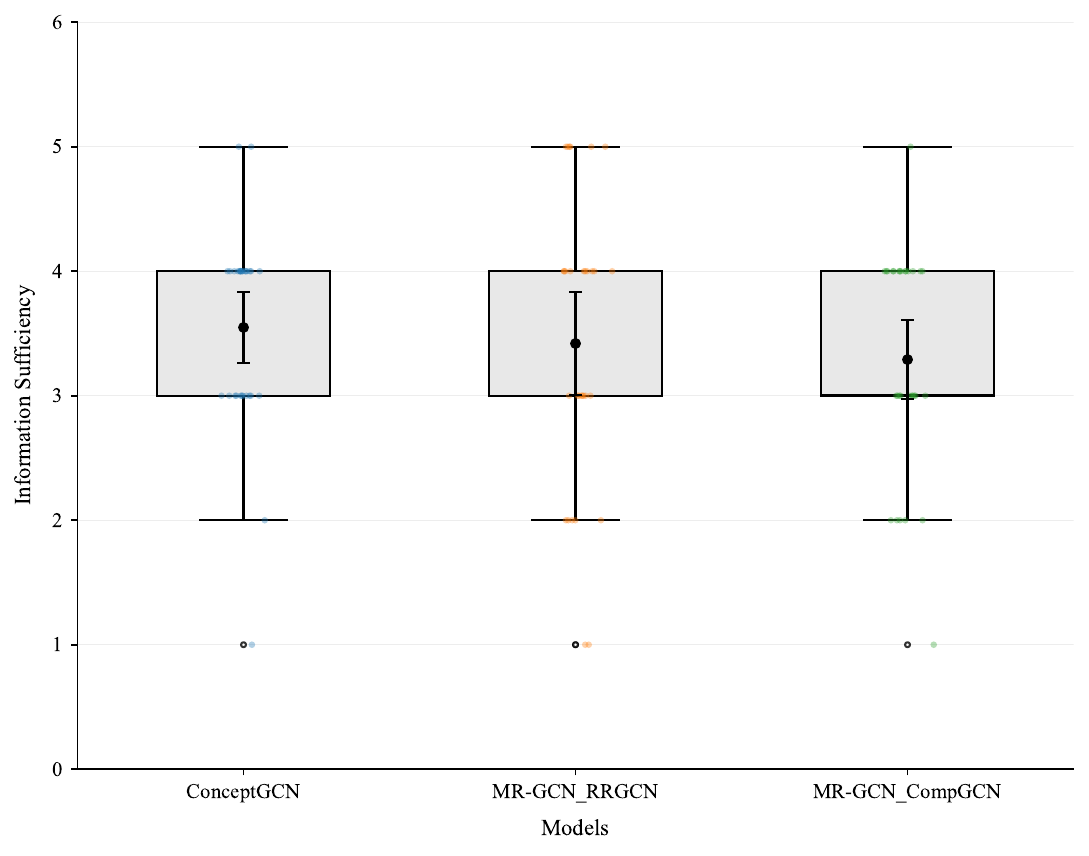}
    \caption*{Information Sufficiency}
\end{subfigure}
\vspace{0.05cm}
{\footnotesize (a) Perceived System Qualities}
\vspace{0.30cm}
\begin{subfigure}{0.45\linewidth}
    \centering
    \includegraphics[width=\linewidth]{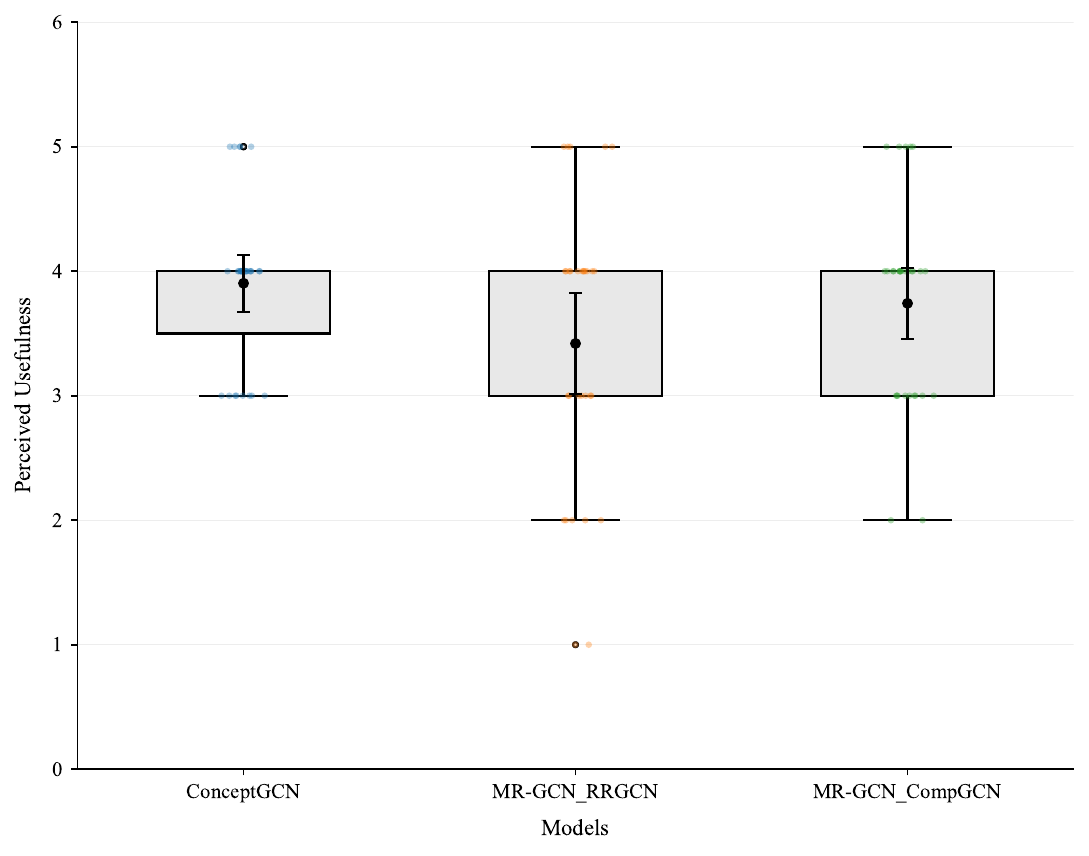}
    \caption*{Perceived Usefulness}
    {\footnotesize (b) Users' Beliefs}
\end{subfigure}
\hfill
\begin{subfigure}{0.45\linewidth}
    \centering
    \includegraphics[width=\linewidth]{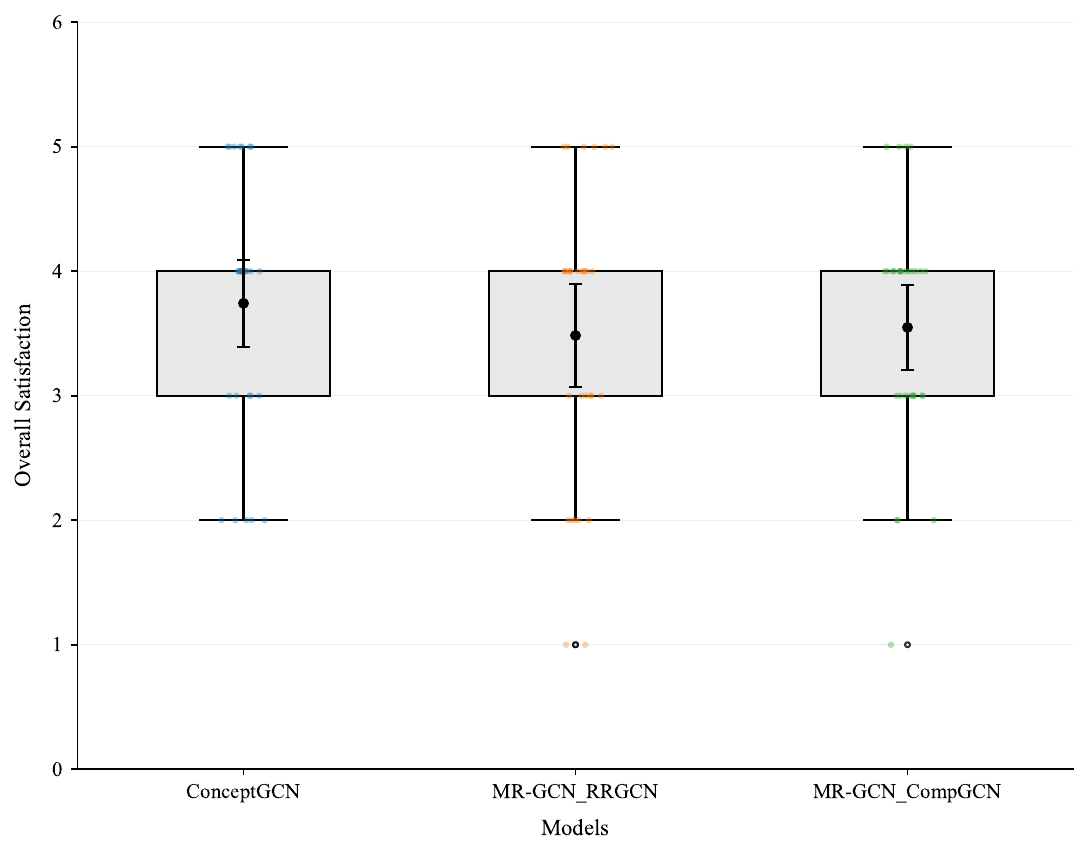}
    \caption*{Overall Satisfaction}
    {\footnotesize (c) Users' Attitudes}
\end{subfigure}
\vspace{0.30cm}
\begin{subfigure}{0.45\linewidth}
    \centering
    \includegraphics[width=\linewidth]{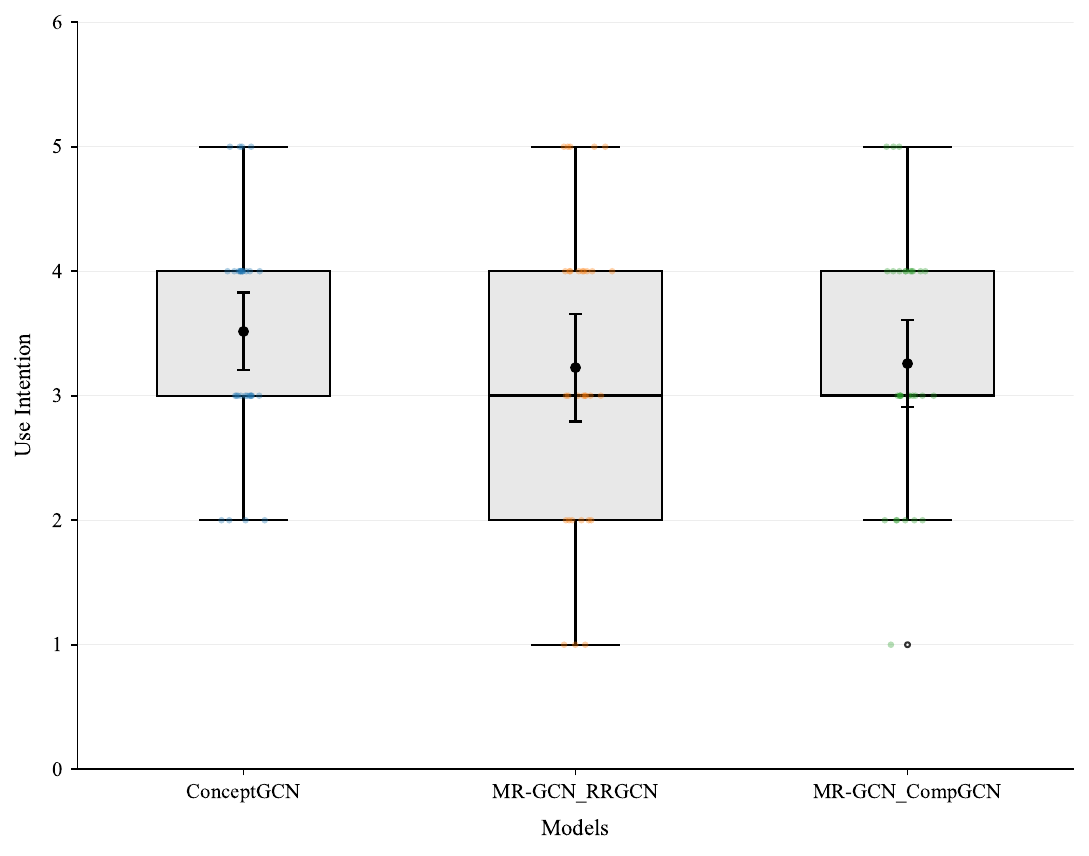}
    \caption*{Use Intention}
    {\footnotesize (d) Behavioral Intentions}
\end{subfigure}
\caption{Evaluation Results Grouped According to the ResQue Framework}
\label{fig:resque_grouped_results}
\end{figure}

To assess users' perceived benefits, all participants evaluated the same set of criteria across the three models. The evaluation results and comparisons between the models across each criterion are presented in Table ~\ref{tab:combined_results} and Figure \ref{fig:resque_grouped_results}.
The Friedman test results are reported using the chi-square statistic $\chi^2$, the $p$-value, and Kendall’s coefficient of concordance ($W$) as the effect size measure.
Overall, the ResQue evaluation results indicate that all criteria scored high for the three models, demonstrating the benefits of a recommendation system based on our proposed sequential learner modeling approach. Moreover,
the results indicate no statistically significant differences between the three models across any of the evaluated criteria, as shown in Table \ref{tab:friedman_precision_constructs}. Specifically, no significant differences were observed for perceived accuracy ($\chi^2(2) = 3.88$, $p = .144$, $W = .06$), diversity ($\chi^2(2) = 0.93$, $p = .627$, $W = .02$), information sufficiency ($\chi^2(2) = 3.03$, $p = .220$, $W = .05$), perceived usefulness ($\chi^2(2) = 5.26$, $p = .072$, $W = .08$), overall satisfaction ($\chi^2(2) = 3.04$, $p = .219$, $W = .05$), and use intention ($\chi^2(2) = 2.47$, $p = .291$, $W = .04$). The effect sizes, as measured by Kendall's W, were consistently small across all criteria, indicating a low magnitude of differences between models. These findings suggest that the three models exhibit comparable performance across all evaluated criteria. This indicates that whether or not to take different relation types into account when enhancing the concept embeddings does not critically impact the perceived performance of the recommender system.

Similar to the results related to system accuracy, participants
perceived the recommendation accuracy high. Concretely, ConceptGCN obtained the highest rating ($M = 3.87$, $SD = 0.81$), followed by MR-GCN\_CompGCN ($M = 3.65$, $SD = 0.80$) and MR-GCN\_RRGCN ($M = 3.45$, $SD = 1.03$). A relatively high number of participants found that the recommended concepts matched their interests. This indicates that the recommender system has well understood the users’ preferences and that the underlying sequential learner model is accurate in capturing users' interests. 

Regarding perceived usefulness, it is particularly worth noting that a relatively high number of participants found that the provided recommendations were useful to them. ConceptGCN achieved the highest rating ($M = 3.90$, $SD = 0.65$), followed by MR-GCN\_CompGCN ($M = 3.74$, $SD = 0.82$) and MR-GCN\_RRGCN ($M = 3.42$, $SD = 1.15$). Perceived usefulness of a recommender system is the extent to which a user finds that using the system would improve their performance, compared with their experiences without its help \citep{puUsercentricEvaluationFramework2011}. The fact that the majority of participants agreed that the recommender system gave them good suggestions reflects the effectiveness of the underlying sequential learner model. 

A similar pattern is observed for recommendation diversity, where ConceptGCN ranked highest ($M = 3.74$, $SD = 1.03$), followed by MR-GCN\_CompGCN ($M = 3.65$, $SD = 1.08$) and MR-GCN\_RRGCN ($M = 3.55$, $SD = 1.18$). A large number of participants found the recommended concepts to be diverse. We posit that the reason for this result was that taking learner’s short-term interests into account has ensured that concepts from different categories were recommended. Another reason we deem responsible for
this result is that DNU concepts were gathered by examining multiple slides. Consequently,
the generated sequential learner model comprised  more diverse concepts that didn’t necessarily correlate, generating recommendations for concepts that were not very similar to each other, thus leading to higher diversity. 

The ResQue evaluation results further show that participants were in general satisfied with the recommender system. For overall satisfaction, ConceptGCN received the highest rating ($M = 3.74$, $SD = 1.00$), followed by MR-GCN\_CompGCN ($M = 3.55$, $SD = 0.96$) and MR-GCN\_RRGCN ($M = 3.48$, $SD = 1.18$). This result can be attributed to the the usefulness of the recommended concepts, as pointed out by \citet{puUsercentricEvaluationFramework2011} who found that perceived usefulness significantly influences users’ attitudes, including overall satisfaction.    

Compared to accuracy, usefulness, diversity, and overall satisfaction, information sufficiency received relatively lower scores. ConceptGCN achieved the highest score ($M = 3.55$, $SD = 0.81$), followed by MR-GCN\_RRGCN ($M = 3.42$, $SD = 1.18$) and MR-GCN\_CompGCN ($M = 3.29$, $SD = 0.90$). This result shows that many participants found the information provided for the recommended concepts not
sufficient for them to understand their DNU concepts.   
When we asked participants about which information provided for the recommended concepts helped them understand their DNU concepts, most participants mentioned that the Wikipedia abstracts and articles helped them form an initial understanding of the recommended items. However, they also noted that these descriptions were often too general for actual learning purposes and might be outdated, since the update frequency is relatively low.
Regarding the recommended sequences, i.e., the prerequisite relations among concepts, most participants considered them helpful because they provide a clear starting point for learning. However, some participants found them less helpful, mainly because not all recommended concepts were connected through prerequisite relations. As a result, beyond the sequences provided, they still did not know how to approach the isolated recommended concepts.
Answering the question about what additional information do they expect that can help them understand their DNU concepts, participants had varied opinions and thoughts which can be classified into two groups. The first group focuses on providing more learning resources, such as videos, diagrams, and examples. The second group emphasizes the connections between DNU concepts and recommended concepts. Participants in this group provided suggestions like distinguishing different categories by using different colors or explicitly illustrating these connections on the recommendation page.

A comparable trend is observed for use intention, where ConceptGCN achieved the highest score ($M = 3.52$, $SD = 0.89$), while MR-GCN\_CompGCN ($M = 3.26$, $SD = 1.00$) slightly outperformed MR-GCN\_RRGCN ($M = 3.23$, $SD = 1.23$). Similar to information sufficiency, use intention received relatively lower scores. This result can be attributed to the lack of additional information (e.g., videos, examples) provided for the recommended concepts which can help users better understand their DNU concepts. 

The ResQue evaluation results further show that 
ConceptGCN achieved the highest mean scores across all criteria. 
One possible explanation is that the PKG constructed from the learning material used in the study was not sufficiently complex. In particular, it contained relatively few \textit{PREREQUISITE\_TO} relationships compared with \textit{RELATED\_TO} relationships. As a result, the advantages of multi-relational GCN models may not have been fully leveraged. This result may also indicate that, in the recommendation task, the semantic relatedness between concepts plays a more important role than different relation types. However, further experiments are required to validate these interpretations. 
In addition, the ResQue evaluation results reveal that MR-GCN\_CompGCN performed slightly better than MR-GCN\_RRGCN. This indicates that integrating edge embeddings can improve node representations, as edges encode additional relational information. Furthermore, explicitly modeling relationship direction (e.g., input, output) can produce richer and more semantically informative node representations. 

In summary, the findings demonstrate that our proposed approach MR-ConceptGCN for sequential learner modeling is effective in increasing the accuracy of the recommendations and promoting users’ perceptions of the recommender system in terms of perceived
system qualities (perceived accuracy, diversity, information sufficiency), users’ beliefs
(perceived usefulness), users’ attitudes (overall satisfaction), and behavioral intentions
(use intention). 
In particular, the findings indicate that integrating semantic relationships among concepts while jointly modeling users’ long-term and short-term interests positively influences perceived accuracy, usefulness, diversity, and overall satisfaction with the educational recommender system.

\begin{table}[htbp]
\centering
\caption{Descriptive statistics (Mean and Standard Deviation) for all evaluation criteria}
\label{tab:combined_results}
\begin{tabular}{l l c c}
\hline
\textbf{Criterion} & \textbf{Model} & \textbf{M} & \textbf{SD} \\
\hline
\multirow{3}{*}{Perceived Accuracy}
 & ConceptGCN        & 3.87 & 0.81 \\
 & MR-GCN\_RRGCN     & 3.45 & 1.03 \\
 & MR-GCN\_CompGCN   & 3.65 & 0.80 \\
\hline
\multirow{3}{*}{Diversity}
 & ConceptGCN        & 3.74 & 1.03 \\
 & MR-GCN\_RRGCN     & 3.55 & 1.18 \\
 & MR-GCN\_CompGCN   & 3.65 & 1.08 \\
\hline
\multirow{3}{*}{Information Sufficiency}
 & ConceptGCN        & 3.55 & 0.81 \\
 & MR-GCN\_RRGCN     & 3.42 & 1.18 \\
 & MR-GCN\_CompGCN   & 3.29 & 0.90 \\
\hline
\multirow{3}{*}{Perceived Usefulness}
 & ConceptGCN        & 3.90 & 0.65 \\
 & MR-GCN\_RRGCN     & 3.42 & 1.15 \\
 & MR-GCN\_CompGCN   & 3.74 & 0.82 \\
\hline
\multirow{3}{*}{Overall Satisfaction}
 & ConceptGCN        & 3.74 & 1.00 \\
 & MR-GCN\_RRGCN     & 3.48 & 1.18 \\
 & MR-GCN\_CompGCN   & 3.55 & 0.96 \\
\hline
\multirow{3}{*}{Use Intention}
 & ConceptGCN        & 3.52 & 0.89 \\
 & MR-GCN\_RRGCN     & 3.23 & 1.23 \\
 & MR-GCN\_CompGCN   & 3.26 & 1.00 \\
\hline
\end{tabular}
\end{table}
%%%%%%%%%%%%%%%%%%%%%%%%%%%%%%
%%%%%%%%%%%%%%%%%%%%%%%%%%%%%%%%%%%%%%%%
\begin{table}[htbp]
\centering
\caption{Friedman test results and effect sizes}
\label{tab:friedman_precision_constructs}
\begin{tabular}{lcccc}
\toprule
\textbf{Criterion} & $\boldsymbol{\chi^2 (df)}$ & $\mathbf{p}$ & \textbf{Kendall's $W$} \\
\midrule
Perceived Accuracy & 3.88 (2) & .144 & .06 \\
Diversity & 0.93 (2) & .627 & .02 \\
Information Sufficiency & 3.03 (2) & .220 & .05 \\
Perceived Usefulness & 5.26 (2) & .072 & .08 \\
Overall Satisfaction & 3.04 (2) & .219 & .05 \\
Use intention & 2.47 (2) & .291 & .04 \\
\bottomrule
\end{tabular}
\end{table}
%%%%%%%%%%
\section{Limitations}
\label{limitations}
While this research highlights the potential benefits of sequntial learner modeling using multi-relational GCNs, this study is not without
limitations. The sample size was moderate and participants
were drawn from a limited set of geographical regions, which may
limit generalizability. In addition, the study
was conducted in the MOOC platform CourseMapper, within a short-term session setting based on a single learning material. 
While this provided a controlled environment for systematic testing, it remains to be seen whether the findings
generalize to other educational environments, or to longer-term use of different learning materials with more PREREQUISITE\_TO relationships. 
%%%%%%%%%
\section{Conclusion and Future Work}
\label{conclusion}
Personal Knowledge Graphs (PKGs) and Graph Neural Networks (GNNs), particularly Graph Convolutional Networks (GCNs), have gained increasing interest as foundation for user modeling. However, existing approaches do not model heterogeneous relation types in PKGs to learn richer and more informative user models. Additionally, they ignore the
user interaction sequence. To address these issues, in this work we presented MR-ConceptGCN,
a novel fully unsupervised approach focused on concept-based sequential learner modeling using multi-relational GCNs (MR-GCNs). We constructed PKGs
that incorporate multiple relationship types between concepts, namely RELATED TO and PREREQUISITE TO.
Moreover, we presented two unsupervised methods for representation learning of PKG items based on RRGCN and CompGCN. Both methods combine MR-GCNs and SBERT to obtain enhanced relation- and semantic-aware representations of the PKG items. Furthermore, building on the enhanced embeddings of the concepts that a learner did
not understand, we constructed a sequential learner model that captures both long-term and short-term learner interests by combining semantic correlation, sequential order, and positional signals. 
We conducted an online user study
to investigate the impact of an educational recommender system based on MR-ConceptGCN on users’ perceptions of several important
user-centric aspects. The evaluation
results indicate that our approach for sequential learner modeling is particularly effective in improving users' perceptions of accuracy, usefulness, diversity,
and overall satisfaction with the recommender system. 
In future work, we plan to explore richer semantic interaction functions beyond cosine similarity, and more robust handling of noisy or sparse relations. 
Moreover, we will conduct a more extensive user study to compare MR-ConceptGCN with a static learner modeling approach that does not take 
sequential information into account. 
\bibliography{sn-bibliography}

\end{document}